\documentclass{article}

\usepackage[preprint]{corl_2021} 

\usepackage[numbers]{natbib}
\usepackage{multicol}

\usepackage{epsfig}
\usepackage{graphicx}
\usepackage{amsmath}
\usepackage{amssymb}
\usepackage{gensymb}

\usepackage[utf8]{inputenc} 
\usepackage[T1]{fontenc}    
\usepackage{url}            
\usepackage{booktabs}       
\usepackage{amsfonts}       
\usepackage{nicefrac}       
\usepackage{microtype}      

\usepackage{subcaption}
\usepackage{changepage}
\usepackage{epstopdf}
\usepackage[dvipsnames]{xcolor}
\usepackage{colortbl}
\definecolor{LightCyan}{rgb}{0.8,1,1}
\usepackage{mathtools}
\usepackage{amsmath}
\usepackage{setspace}
\usepackage{booktabs}
\usepackage{array}
\usepackage{algorithm}
\usepackage{algorithmic}
\usepackage{multirow}
\usepackage{wrapfig}
\usepackage{bbm}
\usepackage{blindtext}
\usepackage{wrapfig}
\usepackage{float}

\usepackage{pifont}
\newcommand{\cmark}{\ding{51}}
\newcommand{\xmark}{\ding{55}}

\newcommand{\bb}[1]{\mathbf{#1}}

\def\bal#1\eal{\begin{align*}#1\end{align*}}
\newcommand{\R}{\mathbb{R}}
\newcommand{\Z}{\mathbb{Z}}
\newcommand{\N}{\mathbb{N}}
\newcommand{\T}{\intercal}

\newcommand{\removed}[1]{}

\newcommand{\checktext}[1]{\textcolor{black}{#1}}

\newcommand{\presec}{\vspace*{0mm}}
\newcommand{\postsec}{\vspace*{0mm}}
\newcommand{\presubsec}{\vspace*{0mm}}
\newcommand{\postsubsec}{\vspace*{0mm}}

\newcommand{\prepar}{\vspace*{0pt}}

\definecolor{myyellow}{HTML}{F1c548}

\definecolor{relpc_green}{rgb}{.2, .9, .25}
\newcommand{\relpc}[1]{(\textcolor{relpc_green}{#1\%})}
\newcommand{\Fsf}{$F$@.75}
\newcommand{\Fnsf}{$F_n$@0.75}

\begin{document}

\title{RICE: Refining Instance Masks in Cluttered Environments with Graph Neural Networks}

\author{Christopher Xie$^1$ \hspace{4px} Arsalan Mousavian$^2$ \hspace{4px} Yu Xiang$^2$ \hspace{4px} Dieter Fox$^{1,2}$\\
$^1$University of Washington \hspace{6px} $^2$NVIDIA\\
\tt\small \{chrisxie,fox\}@cs.washington.edu \hspace{2px} \{yux,amousavian\}@nvidia.com
}
\maketitle

\begin{abstract}
Segmenting unseen object instances in cluttered environments is an important capability that robots need when functioning in unstructured environments. 
While previous methods have exhibited promising results, they still tend to provide incorrect results in highly cluttered scenes. 
We postulate that a network architecture that encodes relations between objects at a high-level can be beneficial.
Thus, in this work, we propose a novel framework that refines the output of such methods by utilizing a graph-based representation of instance masks. 
We train deep networks capable of sampling smart perturbations to the segmentations, and a graph neural network, which can encode relations between objects, to evaluate the perturbed segmentations. 
Our proposed method is orthogonal to previous works and achieves state-of-the-art performance when combined with them. 
We demonstrate an application that uses uncertainty estimates generated by our method to guide a manipulator, leading to efficient understanding of cluttered scenes.
Code, models, and video can be found at \url{https://github.com/chrisdxie/rice}.
\end{abstract}


\begin{figure}[h]
\begin{center}
\includegraphics[width=\linewidth]{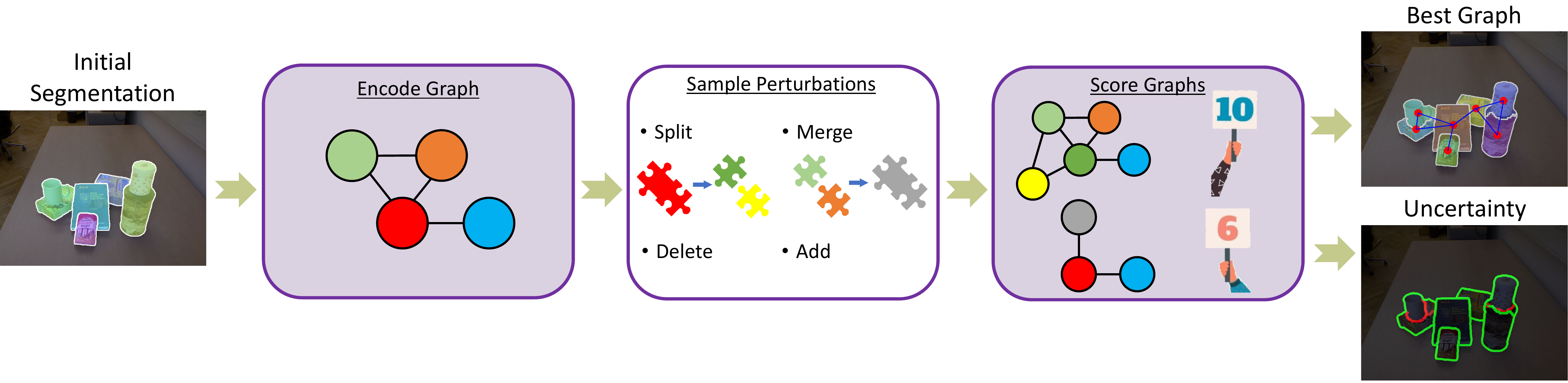}
\caption{High-level overview of our proposed method. Given an initial segmentation, we encode it as a graph, sample perturbations, then score the resulting segmentation graphs. The highest scoring graph and/or contour uncertainty is output. Best viewed in color and zoomed in.}
\label{fig:system_overview}
\end{center}
\end{figure}

\section{Introduction}
\presec

Perception lies at the core of the ability of a robot to function in an unstructured environment. A critical component of such a perception system is its capability to solve Unseen Object Instance Segmentation (UOIS), as it is infeasible to assume all possible objects have been seen in a training phase. Proper segmentation of these unseen instances can lead a better understanding of the scene, which can then be exploited by algorithms such as manipulation~\cite{mousavian2019grasp,murali2020clutteredgrasping,mitash2020task} and re-arrangement~\cite{danielczuk2020object}. 

Many methods for UOIS directly predict segments from raw sensory input such as RGB and/or depth images. While recent methods have shown strong results for this problem~\cite{xie2021unseen, xiang2020learning, he2017mask, jiang2020pointgroup}, they still tend to fail when dealing with highly cluttered scenes, which are common in manipulation scenarios. A natural thought is that an architecture with relational reasoning can benefit the predictions. For example, it can 
potentially learn to recognize common object configurations (e.g. realizing that one object is stacked on top of another). 
While relational inductive biases have shown to be useful for problems such as scene graph prediction~\cite{xu2017scene,chen2019scene,zellers2018neural}, it remains to be seen whether it can be useful in identifying objects in dense clutter. In this work, we investigate the use of graph neural networks, which can encode relations between objects, for segmenting densely cluttered unseen objects. 



In this paper, we propose a novel method for Refining Instance masks in Cluttered Environments, named RICE. Given an initial instance segmentation of unseen objects, we encode it into a \textit{segmentation graph}, where individual masks are encoded as nodes and connected with edges when they are close in pixel space.
Starting from this initial graph, we build a tree of sampled segmentation graphs by perturbing the leaves in a CEM-style (Cross Entropy Method) framework, where example perturbations include splitting and merging. 
We learn Sampling Operation Networks (SO-Nets) that sample efficient and smart perturbations that generally lead to better segmentations. 
The perturbed segmentation graphs are scored with a graph neural network, denoted Segmentation Graph Scoring Network (SGS-Net). 
Finally, we can return the highest scoring segmentation or compute contour uncertainties, depending on the application. Figure~\ref{fig:system_overview} provides a high-level illustration of our method. 

RICE is able to improve the results of existing techniques to deliver state-of-the-art performance for UOIS. An investigatory analysis reveals that applying SGS-Net on top of the SO-Nets results in more accurate and consistent predictions. In particular, we find that SGS-Net learns to rank segmentation graphs better than SO-Nets alone.
\checktext{Additionally, we provide a proof-of-concept efficient scene understanding application that utilizes uncertainties output by RICE to guide a manipulator.}

In summary, our main contributions are: 
1) We propose a novel framework that utilizes a new graph-based representation of instance segmentation masks in cluttered scenes, where we learn deep networks capable of suggesting smart perturbations and scoring of the graphs.
2) Our method achieves state-of-the-art results for UOIS when combined with previous methods. 
\checktext{3) We demonstrate that uncertainty outputs from our method can be used to perform efficient scene understanding.}


\postsec
\section{Related Work}
\presec

\paragraph{Instance Segmentation}
Traditional methods for 2D instance segmentation include GraphCuts~\cite{felzenszwalb2004efficient}, Connected Components~\cite{trevor2013efficient}, and LCCP~\cite{christoph2014object}. Recently, learning-based approaches have provided more semantic solutions. For example, top-down solutions combine segmentation with object proposals in the form of bounding boxes \cite{he2017mask, li2017fully, Chen_2018_CVPR, kirillov2020pointrend}. Mask R-CNN \cite{he2017mask} predicts a foreground mask for each proposal produced by its region proposal network (RPN). However, when bounding boxes contain multiple objects (e.g. cluttered robot manipulation setups), the true instance mask is ambiguous and these methods struggle. Recently, a few methods have investigated bottom-up methods which assign pixels to object instances \cite{de2017semantic, Neven_2019_CVPR, Novotny_2018_ECCV, shao2018clusternet, kong2018recurrent}. Some examples of this include contrastive losses~\cite{de2017semantic} and unrolling mean shift clustering as a neural network to learn pixel embeddings~\cite{kong2018recurrent}.

\removed{
Most of the afore-mentioned algorithms provide instance masks with category-level semantic labels, which do not generalize to unseen objects in novel categories. One approach to adapting these techniques to unseen objects is to employ ``class-agnostic'' training, which treats all object classes as one foreground category \cite{danielczuk2018segmenting}. One family of methods exploits motion cues with class-agnostic training in order to segment arbitrary moving objects \cite{xie2019object, dave2019towards}. Another family of methods are class-agnostic object proposal algorithms \cite{DeepMask, SharpMask, kuo2019shapemask}. However, these methods will segment everything and require some post-processing method to select the masks of interest. \cite{shao2018motion} jointly estimates instance segmentation masks and rigid scene flow, similar to \cite{byravan2017se3nets,byravan2018se3posenets}. 
}

Most of the afore-mentioned algorithms provide instance masks with category-level semantic labels, which do not generalize to unseen objects in novel categories. Class-agnostic methods~\cite{danielczuk2018segmenting, DeepMask, SharpMask, kuo2019shapemask} and motion segmentation~\cite{xie2019object, dave2019towards, shao2018motion} methods have been investigated for this problem. 
In robotic perception, \citet{xie2019uois} proposed to separate the processing of depth and RGB in order to generalize their method from sim-to-real settings and provide sharp masks. Their follow-on work~\cite{xie2021unseen} proposed a 3D voting method to overcome the limitations of their earlier 2D method. \citet{xiang2020learning} showed that training a network on RGB-D with simulated data and a simple contrastive loss~\cite{de2017semantic} can demonstrate strong results for this problem. While these methods show promise, they are not perfect and still admit mistakes in cluttered scenes, which can hamper downstream robot tasks that rely on such perception. Our method is orthogonal to these works, and is designed to refine their outputs by sampling perturbations to result in better instance segmentations in the cluttered environments.

\prepar
\paragraph{Graph Neural Networks}
Graph neural networks (GNN) in vision and robotics have recently become a useful tool for learning relational representations. 
They have found applications in many standard computer vision tasks such as image classification~\cite{garcia2017few,wang2018zero}, object detection~\cite{hu2018relation}, semantic segmentation~\cite{liang2017interpretable}, and question answering~\cite{santoro2017relational}. GNNs have also been used to perform ``scene graph generation'', which requires predicting not just object detections, but also the relations between the objects~\cite{xu2017scene,chen2019scene,zellers2018neural}. The resulting scene graphs have been used for applications such as image retrieval~\cite{johnson2015image}. 
GNNs have also been used to learn object dynamics, properties, and relations for applications such as differential physics engines~\cite{battaglia2016interaction,chang2016compositional}. Our proposed work represents instance segmentation masks as graphs and utilizes this architecture in order to refine the predicted masks.

\postsec

\section{Method}
\presec

Our method, RICE, is designed to Refine Instance masks of unseen objects in Cluttered Environments. 
Given an initial segmentation mask $S \in \N^{H \times W}$ of unseen objects, we first encode this as a \textit{segmentation graph} $G_S$, which is described in Section~\ref{subsec:node_encoder}. 
Then, in Section~\ref{subsec:build_sample_tree}, we build a tree $T$ of sampled segmentation graphs by perturbing the leaves in a CEM-style~\cite{de2005tutorial} framework.
Section~\ref{subsec:sampling_operations} details the sampling operations, which are parameterized by our Sampling Operation Networks (SO-Nets).
Each candidate graph (tree node) is scored by a GNN named Segmentation Graph Scoring Network (SGS-Net), introduced in Section~\ref{subsec:sgs_net}.
Finally, the highest scoring graph in $T$ and/or contour uncertainties are returned. 
Figure~\ref{fig:system_overview} provides a high-level illustration of RICE, and pseudocode can be found in the Supplement (Algorithm~\ref{alg:sample_tree_CEM}).

\subsection{Node Encoder}
\label{subsec:node_encoder}
\presubsec

\begin{wrapfigure}{r}{0.5\textwidth}
  \centering
  \includegraphics[width=\linewidth]{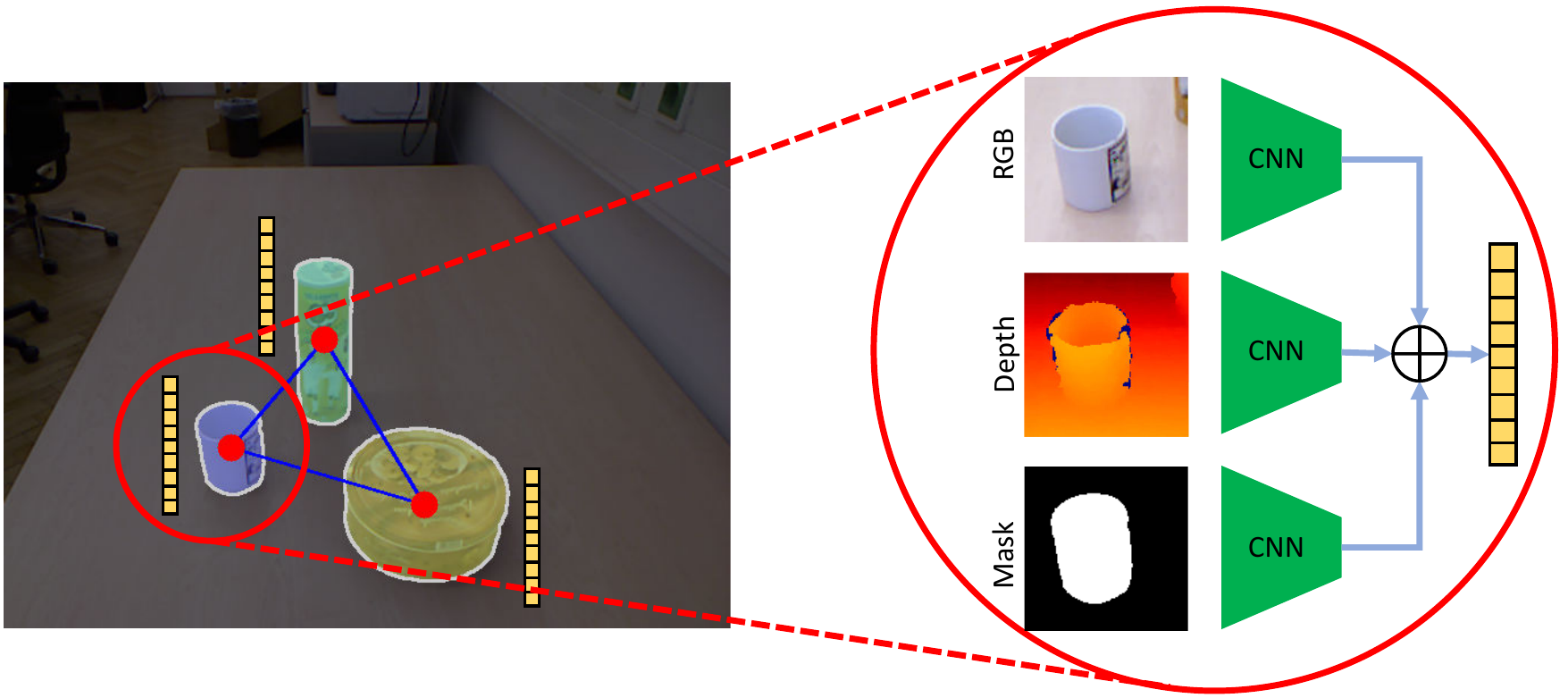}
  \caption{Given an initial instance segmentation mask (left), our segmentation graph representation encodes each individual mask as a graph node (\textcolor{red}{red} dots) with a corresponding feature vector $\bb{v}_i$ (\textcolor{myyellow}{yellow} bar) output by the Node Encoder (right). Edges (\textcolor{blue}{blue} lines) connect nearby masks.}
  \label{fig:node_encoder}
\end{wrapfigure}

Given a single instance mask $S_i \in \{0,1\}^{H \times W}$ for instance $i$, we crop the RGB image $I \in \R^{H \times W \times 3}$, an organized point cloud $D \in \R^{H \times W \times 3}$ (computed by backprojecting a depth image with camera intrinsics), and the mask $S_i$ with some padding for context. 
We then resize the crops to $h \times w$ and feed these into a multi-stream encoder network which we denote as the Node Encoder. This network applies a separate convolutional neural network (CNN) to each input, and then fuses the flattened outputs to provide a feature vector $\bb{v}_i$ for this node. See Figure~\ref{fig:node_encoder} for a visual illustration of the network. Note that we also encode the background mask as a node in the graph.
This gives the segmentation graph $G_S = (V, E)$, where each $\bb{v}_i \in V$ corresponds to an individual instance mask, and nodes are connected with undirected edges $e = (i,j) \in E$ if their set distance is less than a threshold. 

\postsubsec

\subsection{Building the Sample Tree}
\label{subsec:build_sample_tree}
\presubsec

Our sample tree-building procedure operates in a CEM-style fashion. CEM~\cite{de2005tutorial} is an iterative sampling-based optimization algorithm that updates its sampling distribution based on an ``elite set'' of the top $k$ (or top percentile) samples. For more details, we refer the reader to \cite{de2005tutorial}. Following this terminology, our elite set consists of the leaves of our sample tree $T$, each of which are guaranteed to be better with respect to our proxy objective function, SGS-Net. Then, the sampling distribution is implicitly defined by the SO-Nets; while we cannot explicitly write out the distribution, we can certainly sample from it with our sampling operations described in Section~\ref{subsec:sampling_operations}.

Our sample tree $T$ starts off with the root $G_S$. We expand the tree from the leaves with $K$ expansion iterations. For each expansion iteration, we iterate through the current leaves of $T$. For a leaf $G$, we randomly choose a sample operation from Section~\ref{subsec:sampling_operations} and apply it to $G$ to obtain candidate graph $G'$. We then compare the scores $s_G, s_{G'}$ output by SGS-Net, and add $G'$ to $T$ as a child of $G$ if $s_{G'} > s_G$. Thus, any leaf of $T$ is guaranteed to be at least as good as the root $G_S$ w.r.t. our proxy objective function SGS-Net. We apply this procedure $B$ times for $G$, such that each tree node can have a maximum of $B$ children. Thus, $B$ is a branching factor. Finally, due to constraints of limited GPU memory, we exit the process in an anytime fashion whenever we exceed a budget of maximum graph nodes and/or graph edges (not to be confused with tree nodes/edges). See the Supplement for pseudocode (Algorithm~\ref{alg:sample_tree_CEM}) and an example of the sample tree-building procedure (Figure~\ref{fig:sample_tree_example}). 

It is important to note that while we utilize our learned SO-Nets and SGS-Net to build the sample tree $T$, they are applied in different manners (although they are trained on the same dataset). In order to add a candidate graph to the tree, they must both agree in the sense that the perturbation must be suggested via an SO-Net and SGS-Net must approve of the candidate graph via its score. This redundancy offers a level of robustness, Section~\ref{subsec:ablation} shows that the combination of these leads to more accurate performance with lower variance.


\postsubsec

\subsection{Sampling Operations}
\label{subsec:sampling_operations}
\presubsec

\begin{figure*}[t]
     \centering
     \begin{subfigure}[b]{0.24\textwidth}
         \centering
         \includegraphics[width=\textwidth]{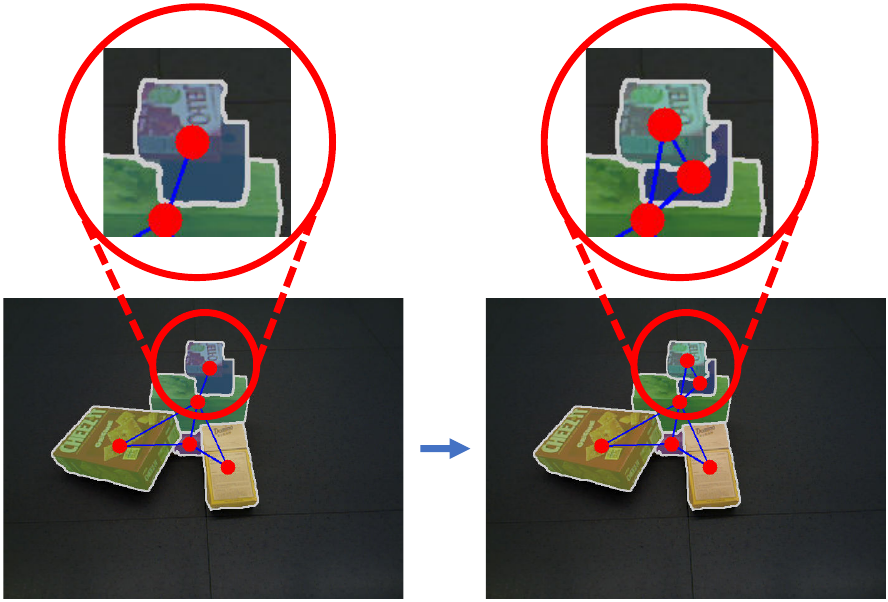}
         \caption{Split}
         \label{fig:split_op}
     \end{subfigure}
     \hfill
     \begin{subfigure}[b]{0.24\textwidth}
         \centering
         \includegraphics[width=\textwidth]{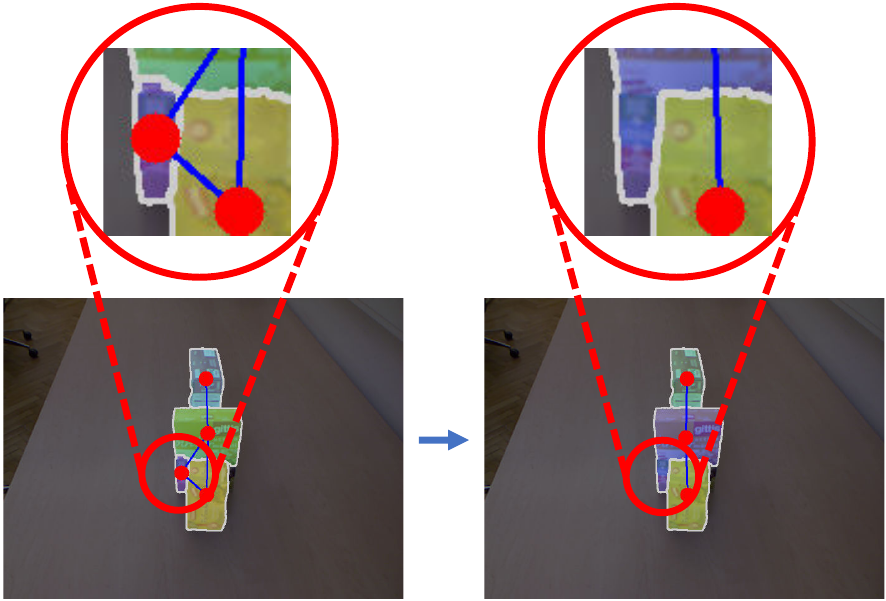}
         \caption{Merge}
         \label{fig:merge_op}
     \end{subfigure}
     \hfill
     \begin{subfigure}[b]{0.24\textwidth}
         \centering
         \includegraphics[width=\textwidth]{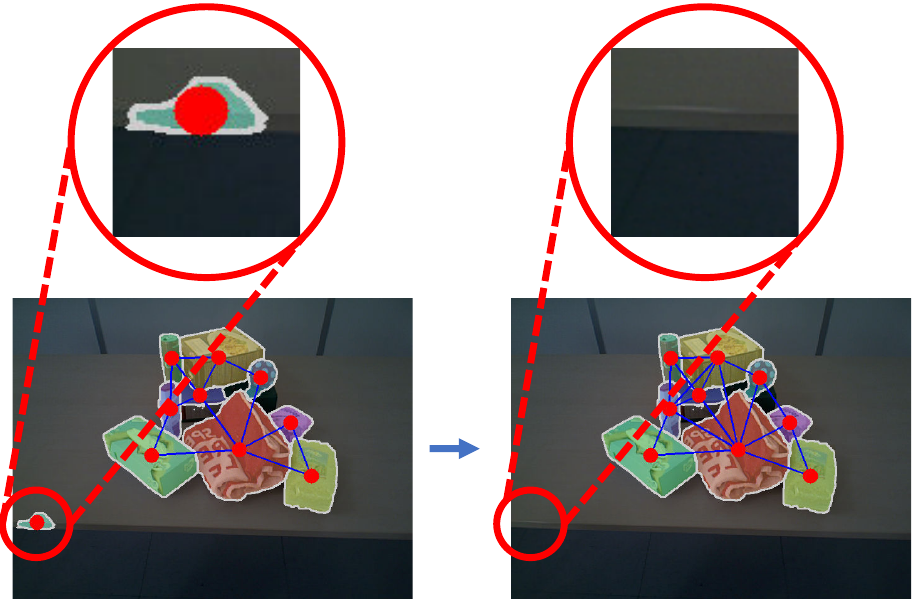}
         \caption{Delete}
         \label{fig:delete_op}
     \end{subfigure}
     \hfill
     \begin{subfigure}[b]{0.24\textwidth}
         \centering
         \includegraphics[width=\textwidth]{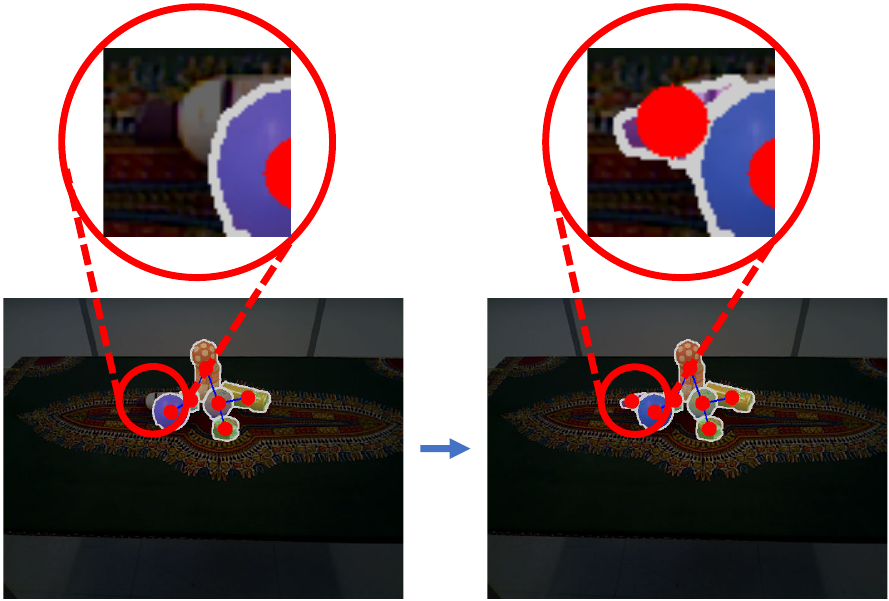}
         \caption{Add}
         \label{fig:add_op}
     \end{subfigure}
        \caption{We show real-world examples of the sampling operations and how they can refine the original segmentation. Best viewed in color on a computer screen and zoomed in.}
        \label{fig:sample_operations}
\end{figure*}

We consider four sampling operations: 1) splitting, 2) merging, 3) deleting, and 4) adding. However, randomly performing these operations leads to inefficient samples which wastes computation time and memory. For example, it is not clear how to split or add an instance mask randomly such that it may potentially result in a better segmentation. Thus, we introduce two networks for these four operations, SplitNet and DeleteNet, which comprise our SO-Nets. They are learned to suggest smart perturbations to bias the sampling towards better graphs, lowering the amount of samples needed in order to favorably refine the segmentation. Examples of each operation can be found in Figure~\ref{fig:sample_operations}.

\prepar
\paragraph{Split}
It is not clear how to randomly split a mask such that it provides an effective split. For example, a naive thing to do is to sample a straight line to split the mask, however in many cases this will not result in a reasonable split (see Figure~\ref{fig:split_op} for an example). Thus, we propose to learn a deep network denoted SplitNet to handle this. SplitNet takes the output of the Node Encoder (before flattening), fuses them with concatenation followed by a convolution, then passes them through a single decoder with skip connections. Essentially it is a multi-stream encoder-decoder U-Net~\cite{ronneberger2015u} architecture, much like Y-Net~\cite{xie2019object}, except that it has three streams for RGB, depth, and the mask.
The output of SplitNet is a pixel-dense probability map $p_i \in [0,1]^{h \times w}$ of split-able object boundaries. To sample a split for instance mask $S_i$, we first sample two end points on the contour of the original mask $S_i$, and calculate the highest probability path from the end points that travels through $p_i$, resulting in a trajectory $\tau = \{(u_t, v_t)\}_{t=1}^{L_i}$ of length $L_i$. We score the split with $s_{\tau} = \frac{1}{L_i} \sum_t p_i[u_t, v_t]  \in \R$, which is the average probability along the sampled path. More details can be found in the Supplement (Section~\ref{subsec:splitnet_sampling}).

\prepar
\paragraph{Merge}
We exploit the fact that merging is the opposite of splitting and adapt SplitNet for this operation. For each pair $(i,j)$ of neighboring masks, we take their union $S_{ij}$ and pass it through SplitNet to get $p_{ij}$. Note that we do not consider merging disjoint masks that may belong to the same instance, which is a limitation of this work. To compute the merge score $m_{ij}$, we first compute the union of the boundaries of $S_i$ and $S_j$, denoted $B_{ij} \in \{0,1\}^{h \times w}$. Then, we calculate the merge score as $m_{ij} = 1 - (p_{ij} \odot B_{ij} / (\bb{1}^\T p_{ij} \bb{1}))$ where $\odot$ is element-wise multiplication, $\bb{1}$ is a vector of ones. This is essentially a weighted average of $B_{ij}$ with weights $p_{ij}$.This score indicates how likely SplitNet thinks $S_i$ and $S_j$ correspond to different objects. Figure~\ref{fig:merge_op} shows an ideal merge operation.

\prepar
\paragraph{Delete}
We design a network, DeleteNet, to provide delete scores $d_i \in \R$ for every instance (graph node) $i$. This network is also built on top of the Node Encoder: it computes the difference $\bb{v}_i - \bb{v}_{bg}$, where $\bb{v}_{bg}$ is the feature vector for the background node output by the Node Encoder. This difference is then provided as input to a multi-layer perceptron (MLP) which outputs a scalar $d_i$. See Figure~\ref{fig:delete_op} for an example of how DeleteNet can help remove false positives from the segmentation.

\prepar
\paragraph{Add}
Similarly to merging, we can exploit the fact that adding is the opposite of deleting. Given a candidate mask $S_{N+1}$ to add to the graph, we can use DeleteNet to compute its delete score $d_{N+1}$. If $d_{N+1}$ is below a threshold, we successfully add the mask to the graph. However, the question remains of how to generate such candidate masks. Given an external foreground mask $F \in \{0,1\}^{H \times W}$ (provided by UOIS-Net-3D~\cite{xie2021unseen}), we run connected components on $F \setminus \{\cup_i S_i\}$, and use the discovered components as potential new masks. A successful addition operation can be seen in Figure~\ref{fig:add_op}.


\postsubsec
\subsection{Segmentation Graph Scoring Network}
\label{subsec:sgs_net}
\presubsec

While our sample operations provide efficient samples that typically lead to better segmentation graphs, they can also suggest samples that worsen the segmentation. Thus, we learn SGS-Net which acts as a proxy for the objective function in the CEM framework. Our proposed SGS-Net learns to score a segmentation graph by considering the fused feature vectors $\bb{v}_i$ in context of their neighboring graph nodes (masks). We posit that this context will aid SGS-Net in predicting whether the perturbations improve the segmentation. For example, it can potentially learn to recognize common object configurations from the training set, and score such configurations higher. 

\begin{figure*}[t]
\begin{center}
\includegraphics[width=\linewidth]{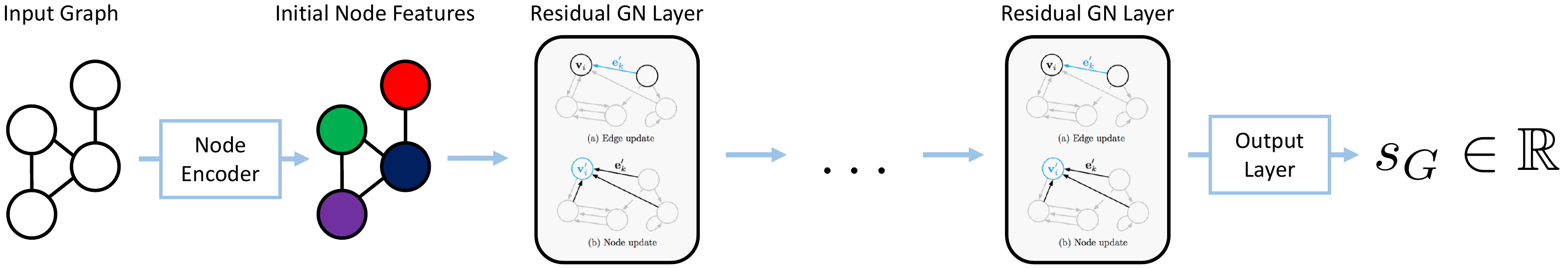}
\caption{A high-level illustration of our Segmentation Graph Scoring Network (SGS-Net). It is composed of a Node Encoder (see Figure~\ref{fig:node_encoder}), multiple Residual GraphNet Layers, and an output layer. We borrowed elements from Figure 3 of \citet{battaglia2018relational}.}
\label{fig:SGS-Net}
\end{center}
\end{figure*}

A high-level illustration of SGS-Net can be found in Figure~\ref{fig:SGS-Net}. The initial node features $\bb{v}_i^{(0)}$ are given by the Node Encoder, and we obtain initial edge features $\bb{e}_{ij}^{(0)}$ by running the Node Encoder on all neighboring union masks $S_{ij}$. Then, we pass them through multiple Residual GraphNet Layers (RGLs), which are essentially GraphNet Layers~\cite{battaglia2018relational} with a residual connection. We refer readers to \citet{battaglia2018relational} for details of GraphNet Layers, and also provide a full mathematical specification of RGLs in the Supplement (Section~\ref{sec:sgs_net_details}) for completeness. The output of SGS-Net is a scalar score in $[0,1]$.

\postsubsec

\subsection{Training Procedure}
\presubsec

For SplitNet, we apply a weighted binary cross entropy (BCE) loss to the probability map $p$: $\ell_{\text{split}} = \sum_u w_u\ \ell_{bce}\left(p_u, \hat{p}_u\right)$ where $u$ ranges over pixels, $\hat{p} \in \{0,1\}^{h\times w}$ is ground truth boundary, and $\ell_{bce}$ is the binary cross entropy loss. The weight $w_u$ is inversely proportional to the number of pixels with labels equal to $\hat{p}_u$.
DeleteNet is also trained with standard BCE loss. 
SGS-Net is trained with $\ell_{bce}$ to regress to $.8 F + .2  F\text{@.75}$, where $F$ is the Overlap F-measure~\cite{xie2019uois} and $F\text{@.75}$ is the Overlap $F$@.75 measure~\cite{ochs2014segmentation}. The latter measures the percentage of correctly segmented instances.
Thus, SGS-Net learns to predict a score based on the number of correctly identified pixels and instances. Note that this regression problem is very difficult to solve. However, the scores do not actually matter as long as the relative scoring is correct, since building the sample tree relies only on this  (Section~\ref{subsec:build_sample_tree}). In Section~\ref{subsec:ranking_experiment} we show that while SGS-Net may not solve the regression problem well, it learns to rank graphs accurately. 
Further training and implementation details can be found in the Supplement (Section~\ref{sec:implementation_details}).

\postsec
\section{Experiments}
\label{sec:experiments}
\presec

\subsection{Encoding RGB and Modality Tuning}
\presubsec

We use ResNet50~\cite{he2016deep} with Feature Pyramid Networks~\cite{lin2017feature} (FPN) to encode RGB images before passing them to the Node Encoder. However, since we are training with (a more cluttered version of) the non-photorealistic synthetic dataset from \citet{xie2019uois}, we perform modality tuning~\cite{aytar2017cross}, where we fine-tune earlier convolutional layers of ResNet50 during training, and use the COCO~\cite{lin2014microsoft} pretrained weights during inference. 
For all experiments, we modality tune the \texttt{conv1} and \texttt{conv2\_1} blocks of ResNet. We provide an experiment in the Supplement (Section~\ref{subsec:modality_tuning}) that shows this setting is optimal.

\postsubsec
\subsection{Datasets and Metrics}
\presubsec

We evaluate our method on two real-world datasets of challenging cluttered tabletop scenes: OCID \cite{suchi2019easylabel} and OSD \cite{richtsfeld2012segmentation}, which have 2346 images of semi-automatically constructed labels and 111 manually labeled images, respectively. Our SO-Nets and SGS-Net are trained on a more cluttered version of the synthetic Tabletop Object Dataset (TOD)~\cite{xie2019uois}, where each scene has anywhere between 20 and 30 ShapeNet~\cite{shapenet2015} objects. We use 20k scenes in total, with 5 images per scene.

\citet{xie2019uois} introduced the Overlap P/R/F and Boundary P/R/F measures for the problem of UOIS. However, these metrics do not weight objects equally; they are dependent on the size and larger objects tend to dominate the metrics. Thus, we introduce a variation to these metrics that equally weights the errors of individual objects regardless of their size. Given a Hungarian assignment $A$ between the predicted instance masks $\{S_i\}_{i=1}^N$ and the ground truth instance masks $\{\hat{S}_j\}_{j=1}^M$, we compute our Object Size Normalized (OSN) P/R/F measures as follows:
\begin{equation*}
    P_n = \frac{\sum \limits_{(i,j) \in A} P_{ij}}{N},\ \ R_n = \frac{\sum \limits_{(i,j) \in A} R_{ij}}{M},\ \ F_n = \frac{\sum \limits_{(i,j) \in A} F_{ij}}{\max(M,N)},\ \ F_n\text{@.75} = \frac{\sum \limits_{(i,j) \in A} \bb{1}\{F_{ij} >= 0.75\}}{\max(M,N)},
\end{equation*}
where $P_{ij}, R_{ij}, F_{ij}$ are the precision, recall, and F-measure of $S_i, \hat{S}_j$. Note that the $F_n\text{@.75}$ penalizes both false positive and false negative instances, as opposed to the normal $F\text{@.75}$, which does not penalize false positives. Similarly to \citet{xie2019uois}, we can apply the OSN metrics to the pixels and boundaries, giving us Overlap and Boundary $P_n/R_n/F_n$ measures. For comparison, we also show results with the normal Overlap and Boundary P/R/F measures in the appendix. 

We run each experiment 5 times and show means and standard deviations for all metrics.


\postsubsec
\subsection{SOTA Improvements}
\label{subsec:sota_improvements}
\presubsec

\begin{figure*}[t]
     \centering
     \begin{subfigure}[b]{0.49\textwidth}
         \centering
         \includegraphics[width=\textwidth]{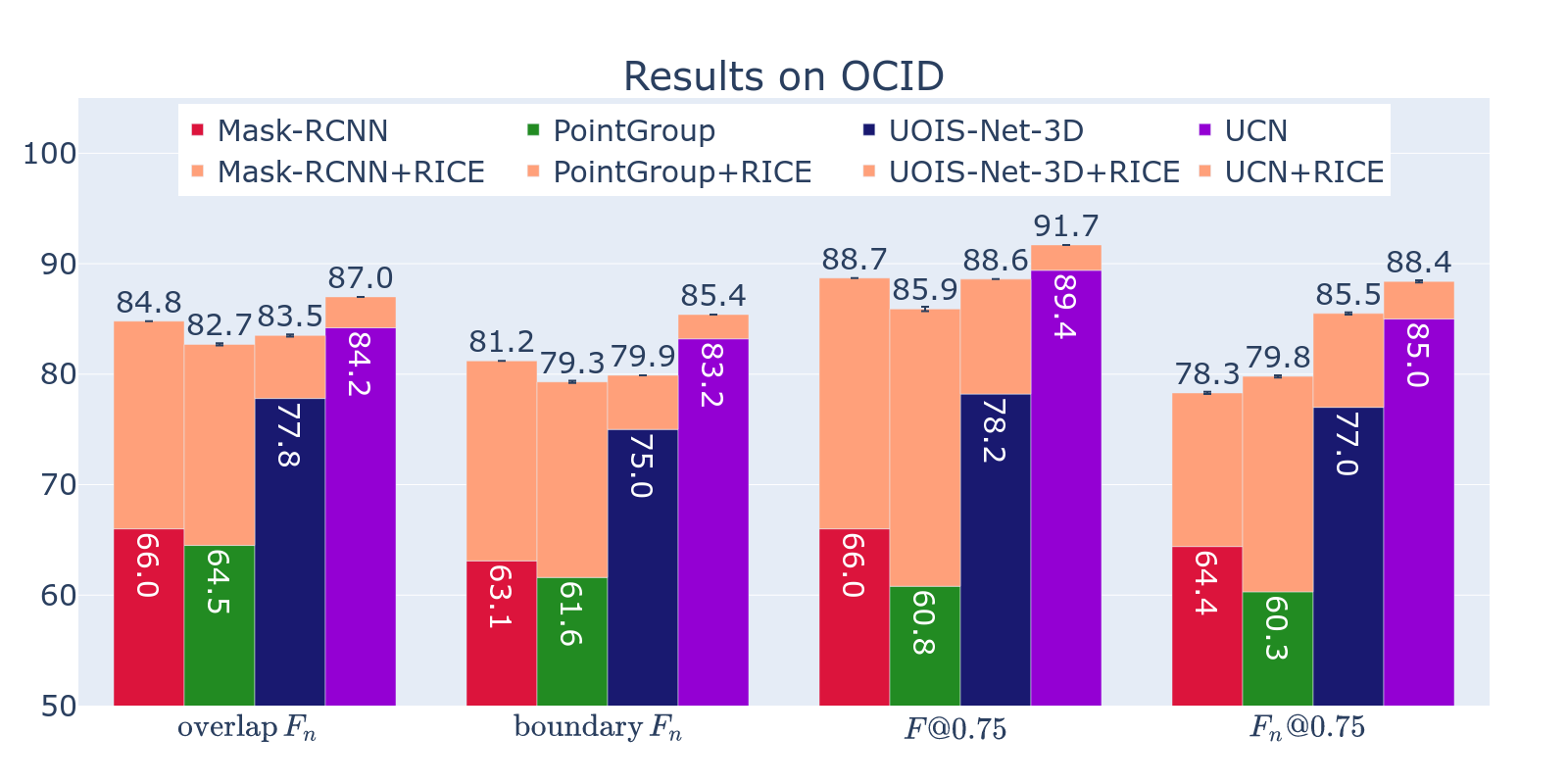}
     \end{subfigure}
     \hfill
     \begin{subfigure}[b]{0.49\textwidth}
         \centering
         \includegraphics[width=\textwidth]{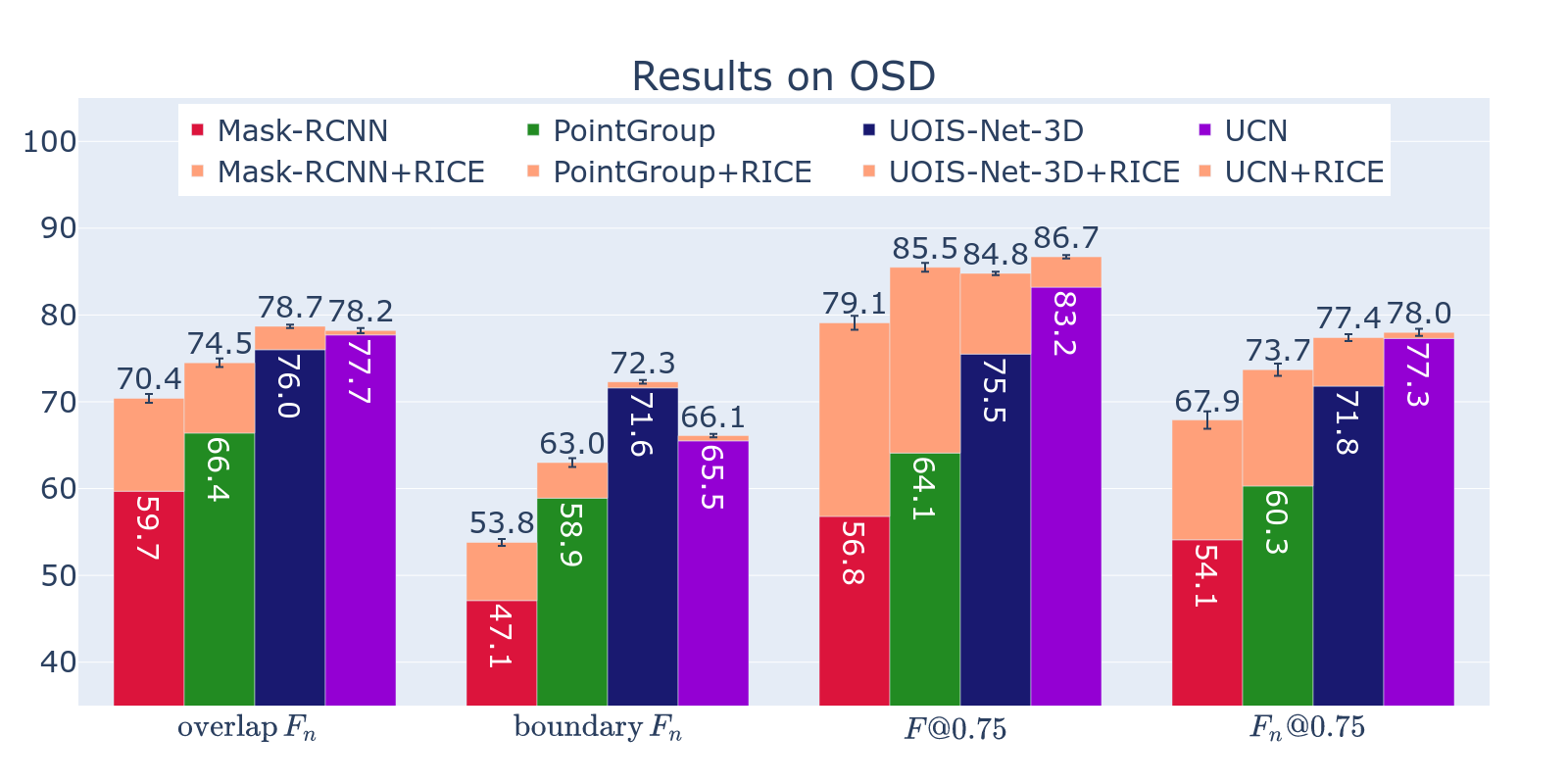}
     \end{subfigure}
     \caption{Applying RICE to refine results from state-of-the-art instance segmentation methods leads to improved performance across the board. Note that standard deviation bars are shown, but are very tight and difficult to see.}
         \label{fig:SOTA_improvements}
\end{figure*}

We demonstrate how RICE can improve upon predicted instance segmentations from state-of-the-art methods. In particular, we apply it to the results of Mask R-CNN~\cite{he2017mask}, PointGroup~\cite{jiang2020pointgroup}, UOIS-Net-3D~\cite{xie2021unseen}, and UCN~\cite{xiang2020learning}. We employ RICE by returning the best segmentation from the leaves as scored by SGS-Net. For brevity, we only show Overlap $F_n$, Boundary $F_n$, $F$@.75, and $F_n$@.75 in Figure~\ref{fig:SOTA_improvements} on both OCID and OSD. The light orange bars show the additional performance that RICE provides over the output of the methods. Standard deviations are shown as error bars, but are in general very narrow, showing that our method provides consistent results despite its stochasticity. RICE provides substantial improvements to all methods. The largest gains occur in Mask R-CNN and PointGroup, with 21.6\% and 32.3\% relative gain in $F_n$@.75 on OCID, respectively. Additionally, on the already strong results from UOIS-Net-3D and UCN, RICE achieves 11.0\% and 4.0\% relative gain in $F_n$@.75 on OCID, respectively. These results are similar on OSD, with the gains being slightly less pronounced, which we believe is due to OSD being a smaller dataset with less clutter. Note that applying RICE increases both $F$@.75 and $F_n$@.75, indicating that not only is it capturing the object identities correctly, it is not simultaneously predicting more instances (false positives). In the appendix, we show full results for all metrics including $P_n,R_n$, and normal P/R/F metrics.

\postsubsec

\subsection{Ablation Study}
\label{subsec:ablation}
\presubsec

\begin{table*}[t]
\centering
\resizebox{\linewidth}{!}{\begin{tabular}{cc||ccc|ccc|cc}
\toprule
\multirow{2}{*}{SO-Nets} & \multirow{2}{*}{SGS-Net} & \multicolumn{3}{c|}{Overlap} & \multicolumn{3}{c|}{Boundary} & & \\
& & \textcolor{orange}{$P_n$} & \textcolor{cyan}{$R_n$} & \textcolor{purple}{$F_n$} & \textcolor{orange}{$P_n$} & \textcolor{cyan}{$R_n$} & \textcolor{purple}{$F_n$} & \textcolor{OliveGreen}{$F$@0.75} & \textcolor{OliveGreen}{$F_n$@0.75} \\
\hline
\xmark & \xmark & 85.1 (--) & 83.0 (--) & 77.8 (--) & 84.6 (--) & 76.5 (--) & 75.0 (--) & 78.2 (--) & 77.0 (--) \\
\cmark & \xmark & 84.7 (1.23) & 89.4 (0.19) & 82.3 (1.09) & 82.7 (1.37) & 82.8 (0.19) & 78.7 (1.10) & 89.0 (0.26) & 84.2 (1.26) \\
\cmark & \cmark & 86.3 (0.03) & 89.1 (0.01) & 83.6 (0.05) & 84.5 (0.04) & 82.5 (0.04) & 80.0 (0.04) & 88.5 (0.02) & 85.5 (0.05) \\ 
\hline
\end{tabular}}
\caption{Ablation to test the utility of SO-Nets and SGS-Net on OCID~\cite{suchi2019easylabel} starting from UOIS-Net-3D~\cite{xie2021unseen} masks. Only using the sample operator networks (SO-Nets) in an iterative sampling scheme already provides an increase in performance, showing that the smart samples are generally improving the initial segmentations. However, the standard deviations (shown in parentheses) are relatively high. Adding in SGS-Net boosts performance while drastically lowering the variance, demonstrating the efficacy of SGS-Net in consistently filtering out bad suggestions by the SO-Nets.}
\label{table:no_gnn_ablation}
\end{table*}

We aim to answer two questions with this study: 1) how good are the samples suggested by our SO-Nets, and 2) to what degree does SGS-Net increase performance and robustness? We study these questions on the larger OCID. 

Since the SO-Nets alone do not provide scores of the perturbed segmentation graphs, we structure our ablation such that this is not needed in order to answer 1). Our SO-Nets are trained to provide smart perturbations that are closer to the ground truth segmentation, so every sample is supposed to be better than the original graph. With this insight, we design an experiment where we run RICE with branch factor $B=1$ and $K=5$ iterations, always add the candidate graph to the tree without consulting SGS-Net, and return the final graph. Essentially, this can be seen as an iterative segmentation graph refinement procedure where the sampled graph should be better than the previous in every iteration. Starting from initial masks provided by UOIS-Net-3D~\cite{xie2021unseen}, we see in Table~\ref{table:no_gnn_ablation} that applying this iterative sampling scheme with SO-Nets only provides better results on almost all metrics than without. However, adding SGS-Net back into the procedure results in better Overlap $F_n$, Boundary $F_n$, and $F_n$@.75, while significantly reducing the standard deviation of the results by two orders of magnitude. This demonstrates that having SGS-Net in RICE delivers not only more accurate performance, but also more robust performance with relatively small variance, which answers 2). Note that $F$@.75 is slightly lower with $F_n$@.75 higher, indicating that SO-Nets are suggesting more samples that better capture the objects, but are suggesting too many instance segments.

\postsubsec
\subsection{SGS-Net Ranking}
\label{subsec:ranking_experiment}
\presubsec

\begin{wrapfigure}{r}{0.4\textwidth}
  \centering
  \includegraphics[width=\linewidth]{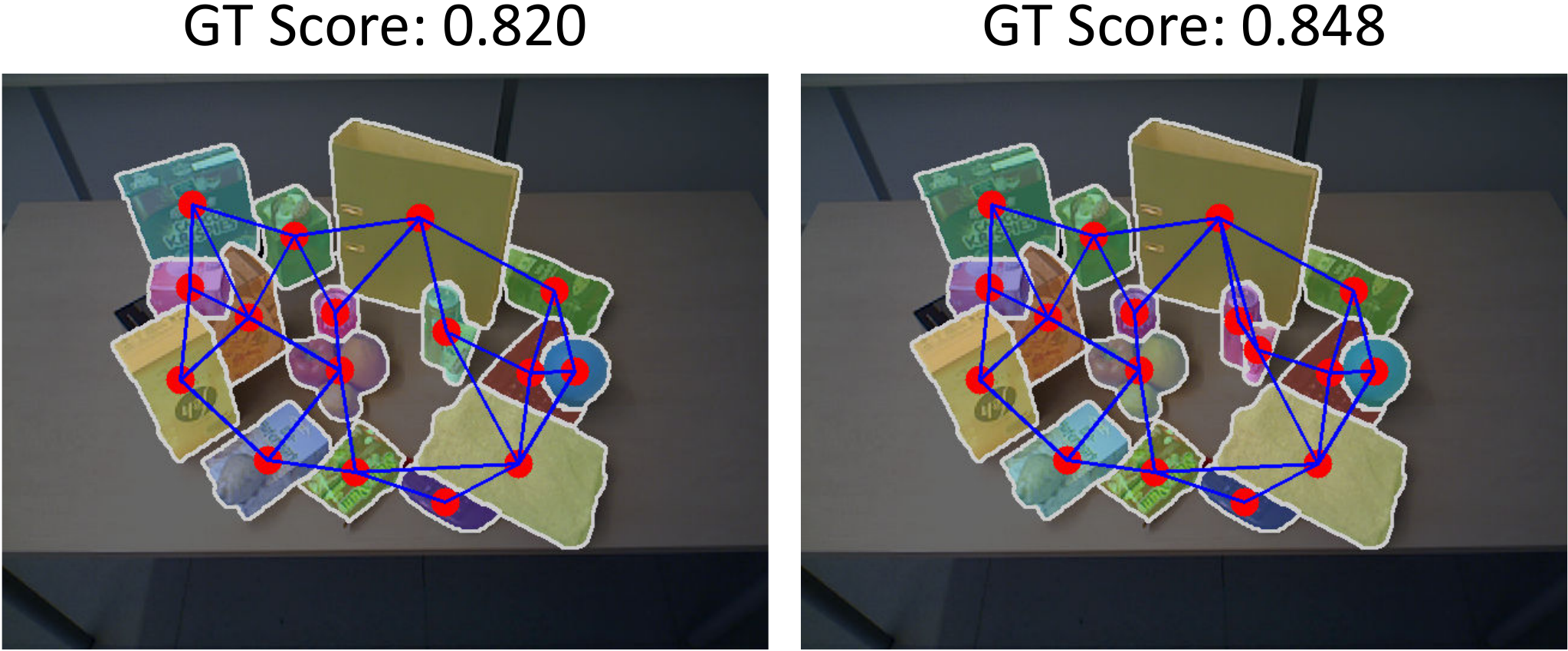}
  \caption{Can you spot the differences between the segmentations?}
  \label{fig:ranking_example}
\end{wrapfigure}

Figure~\ref{fig:ranking_example} shows an example of how difficult scoring the segmentations graphs is; the two slightly different segmentations have a significant difference in their ground truth scores. In fact, SGS-Net does a poor job at scoring the graphs, with a mean absolute error (MAE) of 0.184 and even higher standard deviation shown in Table~\ref{table:gnn_ranking_study}. These values are high given that the scores are in the range $[0,1]$. Then, this begs the question, why does SGS-Net work well within our proposed RICE framework? Recall that the score magnitudes do not matter, only the relative scoring (Section~\ref{subsec:build_sample_tree}). We claim that SGS-Net learns to rank the graphs accurately, and design an experiment to test this hypothesis. 

\begin{wraptable}{R}{0.475\textwidth}
\centering
\begin{tabular}{c||cc}
\toprule
& MAE & nDCG \\ 
\hline
Minimum & -- & 0.844 (0.196) \\
SO-Nets & --  & 0.944 (0.098) \\
SGS-Net & 0.184 (0.212) & 0.952 (0.095) \\
\bottomrule
\end{tabular}
\caption{Ranking study on OCID and OSD.}
\label{table:gnn_ranking_study}
\end{wraptable}

We leverage the normalized Discounted Cumulative Gain (nDCG)~\cite{jarvelin2002cumulated} which is a popular ranking metric in the information retrieval community. The DCG is computed as $\sum_{i=1}^p \frac{2^{\text{rel}_i}-1}{\log_2(i+1)}$ where $\text{rel}_i$ is the numerical relevance of the item at position $i$ (higher is better). This essentially computes a weighted sum of the relevance with a discount factor for further items, which places more emphasis on the high-ranking predictions. The normalized version divides DCG by the ``ideal'' version, i.e. the DCG of the correct ranking. This results in nDCG $\in [0,1]$ with higher being better. We compute nDCG of the ranking of the iterative sampling experiment in Section~\ref{subsec:ablation}, with relevance values in $\{0, ..., K\}$. The ranking for SO-Nets is given by the order of the predicted graphs, and we use SGS-Net scores to compute its ranking/relevance. We also compute the nDCG of the worst ranking, denoted ``minimum''. In Table~\ref{table:gnn_ranking_study}, we see that both SO-Nets and SGS-Net perform significantly better than the worst ranking. SGS-Net provides better ranking than SO-Nets with slightly lower variance, which helps to explain its effectiveness in RICE.

\postsubsec

\subsection{Visualizing Refinements}
\presubsec

\begin{figure*}[t]
\begin{center}
\includegraphics[width=\linewidth]{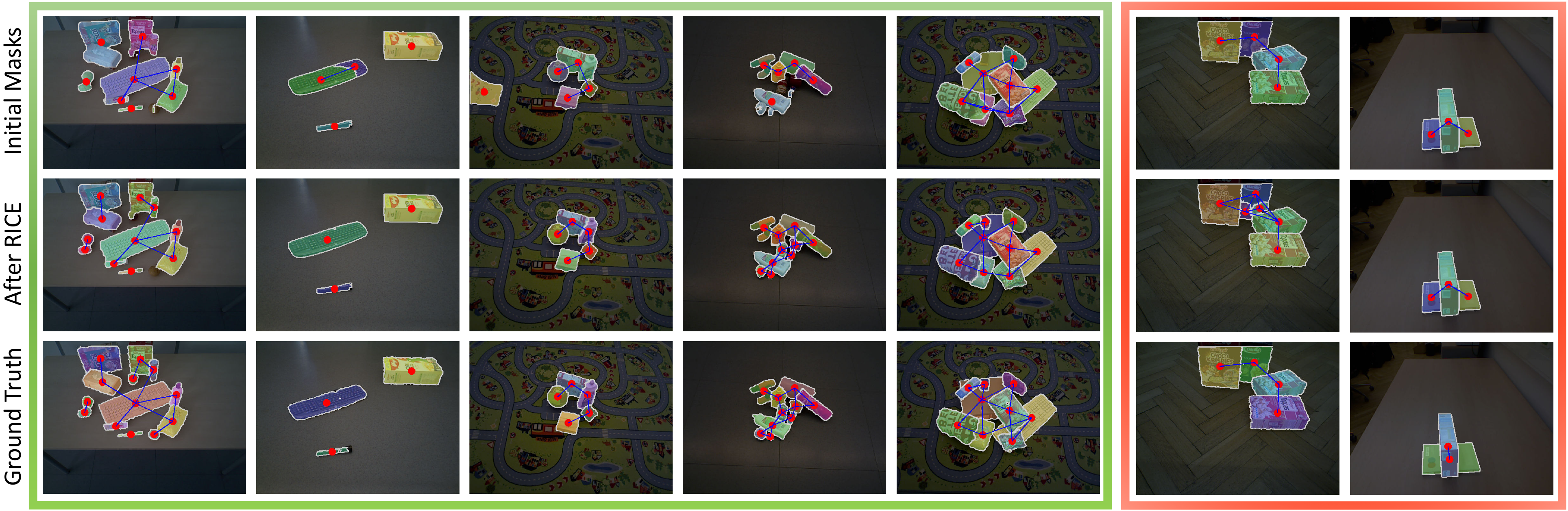}
\caption{We demonstrate successful refinements (left, green box) for each of the sampling operations. Failure modes (right, red box) include textured objects and non-neighboring masks that belong to the same object. Best viewed in color and zoomed in on a computer screen.}
\label{fig:refinements_and_failures}
\end{center}
\end{figure*}

In the left side of Figure~\ref{fig:refinements_and_failures} (green box), we qualitatively demonstrate successful refinements from applying RICE to instance masks provided by state-of-the-art methods. The first column shows an example where many nearby objects are under-segmented. Indeed, RICE manages to find all of the necessary splits except for one. In general, RICE is quite adept at splitting under-segmented instance masks. This is quantitatively confirmed in an additional ablation in the Supplement (Section~\ref{subsec:so_ablation}) that studies the usefulness of each sampling operation. Column two shows an initial mask that is fixed with a merge operation. Column three shows a false positive mask on the textured background, which is suppressed by RICE's deletion sampling operation. In the fourth column, the initial mask is missing quite a few objects, and RICE is able to not only recover them but also correctly segment them, resulting in an almost perfect instance segmentation. In the last column, the bottom left segment is bleeding into a neighboring segment, which is fixed through multiple perturbations (i.e. split, then merge).

\postsubsec
\subsection{Failures and Limitations}
\label{subsec:failures_and_limitations}
\presubsec

In the right side of Figure~\ref{fig:refinements_and_failures} (red box), we discuss some failure modes and limitations. The first column demonstrates a failure mode where RICE tends to over-segment objects with a lot of texture (e.g. cereal box). We believe that this is due to TOD lacking texture on many of its objects~\cite{xie2021unseen}. The second column shows a limitation: since RICE only considers merging neighboring masks, it cannot merge non-neighboring masks that belong to the same object. RICE does nothing and the book is still incorrectly segmented in two pieces. 
We leave this as an interesting avenue for future work.

\postsubsec
\subsection{Guiding a Manipulator with Contour Uncertainties for Efficient Scene Understanding}
\presubsec


\checktext{
Fully segmenting and understanding a scene of cluttered objects is necessary for various manipulation tasks, such as counting objects or re-arranging and sorting them. One way for doing this is to actively singulate each object~\cite{chang2012interactive}. However, such an approach can be extremely inefficient. Here we show how contour uncertainties extracted from RICE can help to solve this problem with potentially far less interactions. Specifically, we extract contour uncertainties by computing the standard deviation of the mask contours of each leaf graph. These uncertainties let us distinguish between objects that are already confidently segmented and those that require physical interaction to resolve segmentation uncertainty. We grasp~\cite{sundermeyer2021icra} any object that has uncertain contours in order to determine its correct segmentation, and repeat this until no more uncertainty persists. Thus, interactions are only required to resolve the uncertain portions of the scene, which can potentially be much less than the number of objects, leading to a more efficient scene understanding method. For example, in Figure~\ref{fig:uncertainty_grasp}, only two grasps are required to fully understand the scene. See the Supplemental video for more results.
}


\begin{figure}[t]
\begin{center}
\includegraphics[width=\linewidth]{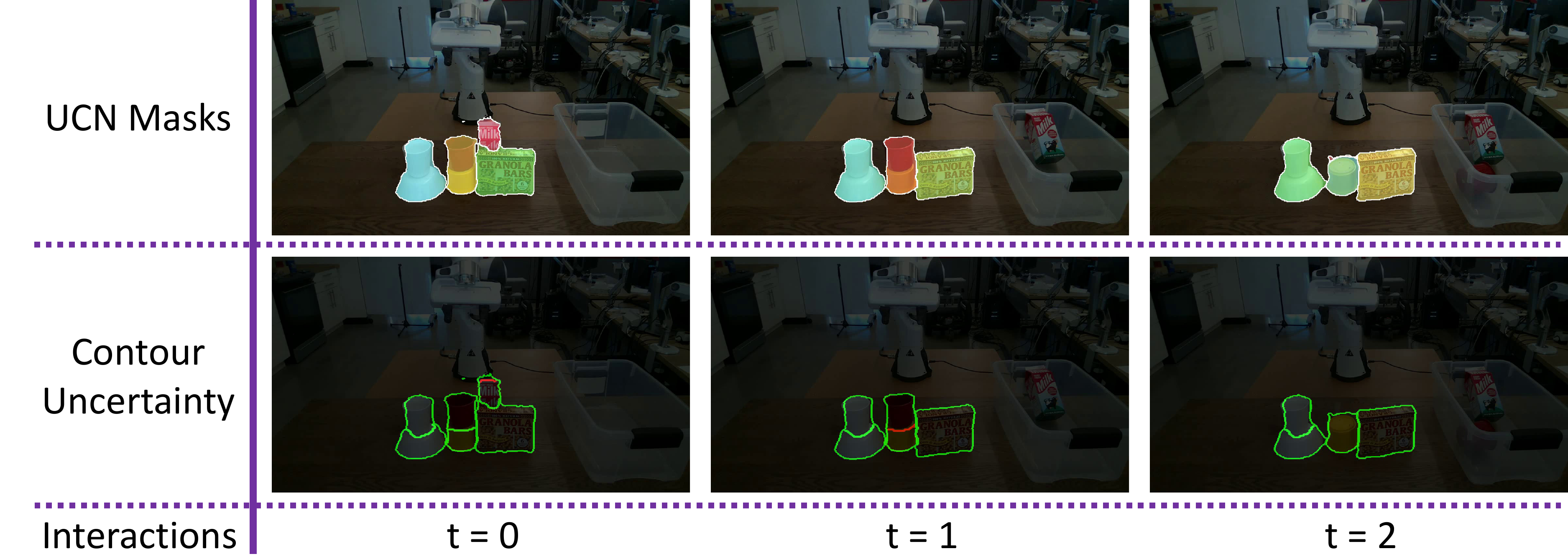}
\caption{\checktext{UCN masks~\cite{xiang2020learning} (top row) and contour uncertainties from RICE (bottom row, uncertainties are shown in red with average contours in green) in a trial of our scene understanding experiment. After grasping the milk carton and red cup, the scene is segmented with full certainty, indicating that the scene is fully understood. Thus, the algorithm terminates without having to singulate each object.}}
\label{fig:uncertainty_grasp}
\end{center}
\end{figure}

\postsec
\section{Conclusion and Future Work}

We have proposed a novel framework that utilizes a graph-based representation of instance segmentation masks. It incorporates deep networks capable of sampling smart perturbations, and a graph neural network that exploits relational inductive biases. Our experimental analysis revealed insight into why our method achieves state-of-the-art performance when combined with previous methods. \checktext{We further demonstrated that our uncertainty outputs can be utilized to perform efficient scene understanding.}

A main limitation of our work is the computational burden; the algorithm runs at 10-15 seconds per frame, depending on the expansion of the sample tree. Additionally, it is GPU-memory intensive as the sample tree must be stored in GPU memory. Future work will explore how to make the method more computationally efficient, along with solving the inherent limitations mentioned in Section~\ref{subsec:failures_and_limitations}.

\bibliography{references}

\begin{thebibliography}{54}
\providecommand{\natexlab}[1]{#1}
\providecommand{\url}[1]{\texttt{#1}}
\expandafter\ifx\csname urlstyle\endcsname\relax
  \providecommand{\doi}[1]{doi: #1}\else
  \providecommand{\doi}{doi: \begingroup \urlstyle{rm}\Url}\fi

\bibitem[Mousavian et~al.(2019)Mousavian, Eppner, and Fox]{mousavian2019grasp}
A.~Mousavian, C.~Eppner, and D.~Fox.
\newblock 6-dof graspnet: Variational grasp generation for object manipulation.
\newblock In \emph{IEEE International Conference on Computer Vision (ICCV)},
  2019.

\bibitem[Murali et~al.(2020)Murali, Mousavian, Eppner, Paxton, and
  Fox]{murali2020clutteredgrasping}
A.~Murali, A.~Mousavian, C.~Eppner, C.~Paxton, and D.~Fox.
\newblock 6-dof grasping for target-driven object manipulation in clutter.
\newblock In \emph{IEEE Conference on Robotics and Automation (ICRA)}, 2020.

\bibitem[Mitash et~al.(2020)Mitash, Shome, Wen, Boularias, and
  Bekris]{mitash2020task}
C.~Mitash, R.~Shome, B.~Wen, A.~Boularias, and K.~Bekris.
\newblock Task-driven perception and manipulation for constrained placement of
  unknown objects.
\newblock \emph{IEEE Robotics and Automation Letters}, 2020.

\bibitem[Danielczuk et~al.(2020)Danielczuk, Mousavian, Eppner, and
  Fox]{danielczuk2020object}
M.~Danielczuk, A.~Mousavian, C.~Eppner, and D.~Fox.
\newblock Object rearrangement using learned implicit collision functions.
\newblock \emph{arXiv preprint arXiv:2011.10726}, 2020.

\bibitem[Xie et~al.(2021)Xie, Xiang, Mousavian, and Fox]{xie2021unseen}
C.~Xie, Y.~Xiang, A.~Mousavian, and D.~Fox.
\newblock Unseen object instance segmentation for robotic environments.
\newblock \emph{IEEE Transactions on Robotics (T-RO)}, 2021.

\bibitem[Xiang et~al.(2020)Xiang, Xie, Mousavian, and Fox]{xiang2020learning}
Y.~Xiang, C.~Xie, A.~Mousavian, and D.~Fox.
\newblock Learning rgb-d feature embeddings for unseen object instance
  segmentation.
\newblock In \emph{Conference on Robot Learning (CoRL)}, 2020.

\bibitem[He et~al.(2017)He, Gkioxari, Doll{\'a}r, and Girshick]{he2017mask}
K.~He, G.~Gkioxari, P.~Doll{\'a}r, and R.~Girshick.
\newblock Mask r-cnn.
\newblock In \emph{IEEE International Conference on Computer Vision (ICCV)},
  2017.

\bibitem[Jiang et~al.(2020)Jiang, Zhao, Shi, Liu, Fu, and
  Jia]{jiang2020pointgroup}
L.~Jiang, H.~Zhao, S.~Shi, S.~Liu, C.-W. Fu, and J.~Jia.
\newblock Pointgroup: Dual-set point grouping for 3d instance segmentation.
\newblock In \emph{IEEE Conference on Computer Vision and Pattern Recognition
  (CVPR)}, 2020.

\bibitem[Xu et~al.(2017)Xu, Zhu, Choy, and Fei-Fei]{xu2017scene}
D.~Xu, Y.~Zhu, C.~B. Choy, and L.~Fei-Fei.
\newblock Scene graph generation by iterative message passing.
\newblock In \emph{Proceedings of the IEEE Conference on Computer Vision and
  Pattern Recognition}, 2017.

\bibitem[Chen et~al.(2019)Chen, Varma, Krishna, Bernstein, Re, and
  Fei-Fei]{chen2019scene}
V.~S. Chen, P.~Varma, R.~Krishna, M.~Bernstein, C.~Re, and L.~Fei-Fei.
\newblock Scene graph prediction with limited labels.
\newblock In \emph{Proceedings of the IEEE International Conference on Computer
  Vision}, 2019.

\bibitem[Zellers et~al.(2018)Zellers, Yatskar, Thomson, and
  Choi]{zellers2018neural}
R.~Zellers, M.~Yatskar, S.~Thomson, and Y.~Choi.
\newblock Neural motifs: Scene graph parsing with global context.
\newblock In \emph{Proceedings of the IEEE Conference on Computer Vision and
  Pattern Recognition}, 2018.

\bibitem[Felzenszwalb and Huttenlocher(2004)]{felzenszwalb2004efficient}
P.~F. Felzenszwalb and D.~P. Huttenlocher.
\newblock Efficient graph-based image segmentation.
\newblock \emph{International Journal of Computer Vision (IJCV)}, 2004.

\bibitem[Trevor et~al.(2013)Trevor, Gedikli, Rusu, and
  Christensen]{trevor2013efficient}
A.~Trevor, S.~Gedikli, R.~Rusu, and H.~Christensen.
\newblock Efficient organized point cloud segmentation with connected
  components.
\newblock In \emph{3rd Workshop on Semantic Perception Mapping and Exploration
  (SPME)}, 2013.

\bibitem[Christoph~Stein et~al.(2014)Christoph~Stein, Schoeler, Papon, and
  Worgotter]{christoph2014object}
S.~Christoph~Stein, M.~Schoeler, J.~Papon, and F.~Worgotter.
\newblock Object partitioning using local convexity.
\newblock In \emph{IEEE Conference on Computer Vision and Pattern Recognition
  (CVPR)}, 2014.

\bibitem[Li et~al.(2017)Li, Qi, Dai, Ji, and Wei]{li2017fully}
Y.~Li, H.~Qi, J.~Dai, X.~Ji, and Y.~Wei.
\newblock Fully convolutional instance-aware semantic segmentation.
\newblock In \emph{IEEE Conference on Computer Vision and Pattern Recognition
  (CVPR)}, 2017.

\bibitem[Chen et~al.(2018)Chen, Hermans, Papandreou, Schroff, Wang, and
  Adam]{Chen_2018_CVPR}
L.-C. Chen, A.~Hermans, G.~Papandreou, F.~Schroff, P.~Wang, and H.~Adam.
\newblock Masklab: Instance segmentation by refining object detection with
  semantic and direction features.
\newblock In \emph{IEEE Conference on Computer Vision and Pattern Recognition
  (CVPR)}, 2018.

\bibitem[Kirillov et~al.(2020)Kirillov, Wu, He, and
  Girshick]{kirillov2020pointrend}
A.~Kirillov, Y.~Wu, K.~He, and R.~Girshick.
\newblock Pointrend: Image segmentation as rendering.
\newblock In \emph{IEEE Conference on Computer Vision and Pattern Recognition
  (CVPR)}, 2020.

\bibitem[De~Brabandere et~al.(2017)De~Brabandere, Neven, and
  Van~Gool]{de2017semantic}
B.~De~Brabandere, D.~Neven, and L.~Van~Gool.
\newblock Semantic instance segmentation with a discriminative loss function.
\newblock \emph{arXiv preprint arXiv:1708.02551}, 2017.

\bibitem[Neven et~al.(2019)Neven, De~Brabandere, Proesmans, and
  Van~Gool]{Neven_2019_CVPR}
D.~Neven, B.~De~Brabandere, M.~Proesmans, and L.~Van~Gool.
\newblock Instance segmentation by jointly optimizing spatial embeddings and
  clustering bandwidth.
\newblock In \emph{IEEE Conference on Computer Vision and Pattern Recognition
  (CVPR)}, 2019.

\bibitem[Novotny et~al.(2018)Novotny, Albanie, Larlus, and
  Vedaldi]{Novotny_2018_ECCV}
D.~Novotny, S.~Albanie, D.~Larlus, and A.~Vedaldi.
\newblock Semi-convolutional operators for instance segmentation.
\newblock In \emph{European Conference on Computer Vision (ECCV)}, 2018.

\bibitem[Shao et~al.(2018)Shao, Tian, and Bohg]{shao2018clusternet}
L.~Shao, Y.~Tian, and J.~Bohg.
\newblock Clusternet: 3d instance segmentation in rgb-d images.
\newblock \emph{arXiv preprint arXiv:1807.08894}, 2018.

\bibitem[Kong and Fowlkes(2018)]{kong2018recurrent}
S.~Kong and C.~Fowlkes.
\newblock Recurrent pixel embedding for instance grouping.
\newblock In \emph{IEEE Conference on Computer Vision and Pattern Recognition
  (CVPR)}, 2018.

\bibitem[Danielczuk et~al.(2019)Danielczuk, Matl, Gupta, Li, Lee, Mahler, and
  Goldberg]{danielczuk2018segmenting}
M.~Danielczuk, M.~Matl, S.~Gupta, A.~Li, A.~Lee, J.~Mahler, and K.~Goldberg.
\newblock Segmenting unknown 3d objects from real depth images using mask r-cnn
  trained on synthetic data.
\newblock In \emph{IEEE Conference on Robotics and Automation (ICRA)}, 2019.

\bibitem[Pinheiro et~al.(2015)Pinheiro, Collobert, and Dollár]{DeepMask}
P.~O. Pinheiro, R.~Collobert, and P.~Dollár.
\newblock Learning to segment object candidates.
\newblock In \emph{Advances in Neural Information Processing Systems
  (NeurIPS)}, 2015.

\bibitem[Pinheiro et~al.(2016)Pinheiro, Lin, Collobert, and Dollár]{SharpMask}
P.~O. Pinheiro, T.-Y. Lin, R.~Collobert, and P.~Dollár.
\newblock Learning to refine object segments.
\newblock In \emph{European Conference on Computer Vision (ECCV)}, 2016.

\bibitem[Kuo et~al.(2019)Kuo, Angelova, Malik, and Lin]{kuo2019shapemask}
W.~Kuo, A.~Angelova, J.~Malik, and T.-Y. Lin.
\newblock Shapemask: Learning to segment novel objects by refining shape
  priors.
\newblock In \emph{IEEE International Conference on Computer Vision (ICCV)},
  2019.

\bibitem[Xie et~al.(2019)Xie, Xiang, Harchaoui, and Fox]{xie2019object}
C.~Xie, Y.~Xiang, Z.~Harchaoui, and D.~Fox.
\newblock Object discovery in videos as foreground motion clustering.
\newblock In \emph{IEEE Conference on Computer Vision and Pattern Recognition
  (CVPR)}, 2019.

\bibitem[Dave et~al.(2019)Dave, Tokmakov, and Ramanan]{dave2019towards}
A.~Dave, P.~Tokmakov, and D.~Ramanan.
\newblock Towards segmenting everything that moves.
\newblock \emph{arXiv preprint arXiv:1902.03715}, 2019.

\bibitem[Shao et~al.(2018)Shao, Shah, Dwaracherla, and Bohg]{shao2018motion}
L.~Shao, P.~Shah, V.~Dwaracherla, and J.~Bohg.
\newblock Motion-based object segmentation based on dense rgb-d scene flow.
\newblock \emph{IEEE Robotics and Automation Letters}, 3:\penalty0 3797--3804,
  2018.

\bibitem[Xie et~al.(2019)Xie, Xiang, Mousavian, and Fox]{xie2019uois}
C.~Xie, Y.~Xiang, A.~Mousavian, and D.~Fox.
\newblock The best of both modes: Separately leveraging rgb and depth for
  unseen object instance segmentation.
\newblock In \emph{Conference on Robot Learning (CoRL)}, 2019.

\bibitem[Garcia and Bruna(2018)]{garcia2017few}
V.~Garcia and J.~Bruna.
\newblock Few-shot learning with graph neural networks.
\newblock In \emph{International Conference on Learning Representations
  (ICLR)}, 2018.

\bibitem[Wang et~al.(2018)Wang, Ye, and Gupta]{wang2018zero}
X.~Wang, Y.~Ye, and A.~Gupta.
\newblock Zero-shot recognition via semantic embeddings and knowledge graphs.
\newblock In \emph{Proceedings of the IEEE conference on computer vision and
  pattern recognition}, 2018.

\bibitem[Hu et~al.(2018)Hu, Gu, Zhang, Dai, and Wei]{hu2018relation}
H.~Hu, J.~Gu, Z.~Zhang, J.~Dai, and Y.~Wei.
\newblock Relation networks for object detection.
\newblock In \emph{Proceedings of the IEEE conference on computer vision and
  pattern recognition}, 2018.

\bibitem[Liang et~al.(2017)Liang, Lin, Shen, Feng, Yan, and
  Xing]{liang2017interpretable}
X.~Liang, L.~Lin, X.~Shen, J.~Feng, S.~Yan, and E.~P. Xing.
\newblock Interpretable structure-evolving lstm.
\newblock In \emph{Proceedings of the IEEE conference on computer vision and
  pattern recognition}, 2017.

\bibitem[Santoro et~al.(2017)Santoro, Raposo, Barrett, Malinowski, Pascanu,
  Battaglia, and Lillicrap]{santoro2017relational}
A.~Santoro, D.~Raposo, D.~G. Barrett, M.~Malinowski, R.~Pascanu, P.~Battaglia,
  and T.~Lillicrap.
\newblock A simple neural network module for relational reasoning.
\newblock In \emph{Advances in Neural Information Processing Systems
  (NeurIPS)}, 2017.

\bibitem[Johnson et~al.(2015)Johnson, Krishna, Stark, Li, Shamma, Bernstein,
  and Fei-Fei]{johnson2015image}
J.~Johnson, R.~Krishna, M.~Stark, L.-J. Li, D.~Shamma, M.~Bernstein, and
  L.~Fei-Fei.
\newblock Image retrieval using scene graphs.
\newblock In \emph{Proceedings of the IEEE conference on computer vision and
  pattern recognition}, 2015.

\bibitem[Battaglia et~al.(2016)Battaglia, Pascanu, Lai, Jimenez~Rezende,
  et~al.]{battaglia2016interaction}
P.~Battaglia, R.~Pascanu, M.~Lai, D.~Jimenez~Rezende, et~al.
\newblock Interaction networks for learning about objects, relations and
  physics.
\newblock \emph{Advances in neural information processing systems},
  29:\penalty0 4502--4510, 2016.

\bibitem[Chang et~al.(2017)Chang, Ullman, Torralba, and
  Tenenbaum]{chang2016compositional}
M.~B. Chang, T.~Ullman, A.~Torralba, and J.~B. Tenenbaum.
\newblock A compositional object-based approach to learning physical dynamics.
\newblock In \emph{International Conference on Learning Representations,
  (ICLR)}, 2017.

\bibitem[De~Boer et~al.(2005)De~Boer, Kroese, Mannor, and
  Rubinstein]{de2005tutorial}
P.-T. De~Boer, D.~P. Kroese, S.~Mannor, and R.~Y. Rubinstein.
\newblock A tutorial on the cross-entropy method.
\newblock \emph{Annals of operations research}, 134\penalty0 (1):\penalty0
  19--67, 2005.

\bibitem[Ronneberger et~al.(2015)Ronneberger, Fischer, and
  Brox]{ronneberger2015u}
O.~Ronneberger, P.~Fischer, and T.~Brox.
\newblock U-net: Convolutional networks for biomedical image segmentation.
\newblock In \emph{International Conference on Medical Image Computing and
  Computer-Assisted Intervention (MICCAI)}, 2015.

\bibitem[Battaglia et~al.(2018)Battaglia, Hamrick, Bapst, Sanchez-Gonzalez,
  Zambaldi, Malinowski, Tacchetti, Raposo, Santoro, Faulkner,
  et~al.]{battaglia2018relational}
P.~W. Battaglia, J.~B. Hamrick, V.~Bapst, A.~Sanchez-Gonzalez, V.~Zambaldi,
  M.~Malinowski, A.~Tacchetti, D.~Raposo, A.~Santoro, R.~Faulkner, et~al.
\newblock Relational inductive biases, deep learning, and graph networks.
\newblock \emph{arXiv preprint arXiv:1806.01261}, 2018.

\bibitem[Ochs et~al.(2014)Ochs, Malik, and Brox]{ochs2014segmentation}
P.~Ochs, J.~Malik, and T.~Brox.
\newblock Segmentation of moving objects by long term video analysis.
\newblock \emph{IEEE transactions on pattern analysis and machine
  intelligence}, 2014.

\bibitem[He et~al.(2016)He, Zhang, Ren, and Sun]{he2016deep}
K.~He, X.~Zhang, S.~Ren, and J.~Sun.
\newblock Deep residual learning for image recognition.
\newblock In \emph{IEEE Conference on Computer Vision and Pattern Recognition
  (CVPR)}, 2016.

\bibitem[Lin et~al.(2017)Lin, Doll{\'a}r, Girshick, He, Hariharan, and
  Belongie]{lin2017feature}
T.-Y. Lin, P.~Doll{\'a}r, R.~Girshick, K.~He, B.~Hariharan, and S.~Belongie.
\newblock Feature pyramid networks for object detection.
\newblock In \emph{IEEE Conference on Computer Vision and Pattern Recognition
  (CVPR)}, 2017.

\bibitem[Aytar et~al.(2017)Aytar, Castrejon, Vondrick, Pirsiavash, and
  Torralba]{aytar2017cross}
Y.~Aytar, L.~Castrejon, C.~Vondrick, H.~Pirsiavash, and A.~Torralba.
\newblock Cross-modal scene networks.
\newblock \emph{IEEE transactions on pattern analysis and machine
  intelligence}, 40\penalty0 (10):\penalty0 2303--2314, 2017.

\bibitem[Lin et~al.(2014)Lin, Maire, Belongie, Hays, Perona, Ramanan,
  Doll{\'a}r, and Zitnick]{lin2014microsoft}
T.-Y. Lin, M.~Maire, S.~Belongie, J.~Hays, P.~Perona, D.~Ramanan,
  P.~Doll{\'a}r, and C.~L. Zitnick.
\newblock Microsoft coco: Common objects in context.
\newblock In \emph{European Conference Computer Vision (ECCV)}, 2014.

\bibitem[Suchi et~al.(2019)Suchi, Patten, and Vincze]{suchi2019easylabel}
M.~Suchi, T.~Patten, and M.~Vincze.
\newblock Easylabel: A semi-automatic pixel-wise object annotation tool for
  creating robotic rgb-d datasets.
\newblock In \emph{IEEE Conference on Robotics and Automation (ICRA)}, 2019.

\bibitem[Richtsfeld et~al.(2012)Richtsfeld, M{\"o}rwald, Prankl, Zillich, and
  Vincze]{richtsfeld2012segmentation}
A.~Richtsfeld, T.~M{\"o}rwald, J.~Prankl, M.~Zillich, and M.~Vincze.
\newblock Segmentation of unknown objects in indoor environments.
\newblock In \emph{IEEE/RSJ International Conference on Intelligent Robots and
  Systems (IROS)}, 2012.

\bibitem[Chang et~al.(2015)Chang, Funkhouser, Guibas, Hanrahan, Huang, Li,
  Savarese, Savva, Song, Su, Xiao, Yi, and Yu]{shapenet2015}
A.~X. Chang, T.~Funkhouser, L.~Guibas, P.~Hanrahan, Q.~Huang, Z.~Li,
  S.~Savarese, M.~Savva, S.~Song, H.~Su, J.~Xiao, L.~Yi, and F.~Yu.
\newblock {ShapeNet: An Information-Rich 3D Model Repository}.
\newblock Technical Report arXiv:1512.03012, 2015.

\bibitem[J{\"a}rvelin and Kek{\"a}l{\"a}inen(2002)]{jarvelin2002cumulated}
K.~J{\"a}rvelin and J.~Kek{\"a}l{\"a}inen.
\newblock Cumulated gain-based evaluation of ir techniques.
\newblock \emph{ACM Transactions on Information Systems (TOIS)}, 20\penalty0
  (4):\penalty0 422--446, 2002.

\bibitem[Chang et~al.(2012)Chang, Smith, and Fox]{chang2012interactive}
L.~Chang, J.~R. Smith, and D.~Fox.
\newblock Interactive singulation of objects from a pile.
\newblock In \emph{2012 IEEE International Conference on Robotics and
  Automation}, 2012.

\bibitem[Sundermeyer et~al.(2021)Sundermeyer, Mousavian, Triebel, and
  Dieter]{sundermeyer2021icra}
M.~Sundermeyer, A.~Mousavian, R.~Triebel, and F.~Dieter.
\newblock Contact-graspnet: Efficient 6-dof grasp generation in
  clutteredscenes.
\newblock In \emph{IEEE International Conference on Robotics and Automation
  (ICRA)}, 2021.

\bibitem[Wu and He(2018)]{wu2018group}
Y.~Wu and K.~He.
\newblock Group normalization.
\newblock In \emph{European Conference on Computer Vision (ECCV)}, 2018.

\bibitem[Kingma and Ba(2015)]{kingma2015adam}
D.~P. Kingma and J.~Ba.
\newblock Adam: A method for stochastic optimization.
\newblock In \emph{International Conference on Learning Representations,
  (ICLR)}, 2015.

\end{thebibliography}

\newpage
\appendix

\section{Sampling Operation Networks details}

\subsection{Architecture Details}

\paragraph{Node Encoder}
The Node Encoder consists of three separate CNN encoders, which consume RGB features (output by Resnet50+FPN~\cite{he2016deep,lin2017feature}), a backprojected XYZ point cloud (from a depth map and known camera intrinsics), and the mask, respectively. Each CNN encoder has 3 blocks of 2 3x3 convolutions followed by a 2x2 max pooling for resolution reduction. Lastly, there is a 7$^\text{th}$ convolution layer. Each convolution is immediately followed by a GroupNorm layer~\cite{wu2018group} and ReLU. The resulting features are 2x2 averaged pooled, flattened, then put through an MLP with 2 hidden layers of dimension 1024 and 512 and an output dimension of 128 (except for SplitNet, which directly consumes the fully convolutional output after the 7$^\text{th}$ conv layer).

\paragraph{SplitNet}
SplitNet builds off of the multi-stream CNN encoders from the Node Encoder. Given the outputs after the 7$^\text{th}$ conv layer of the Node Enocder, these are then concatenated and passed through another conv layer for fusion. Next, this fused output is passed to a U-Net~\cite{ronneberger2015u} style decoder. Skip connections are fed from the multi-stream CNN encoders to the decoder. Essentially, the overall architecture (including the Node Encoder) is a multi-stream encoder-decoder UNet architecture. This is similar to Y-Net~\cite{xie2019object}, except that it has three streams instead of two. Weights of the Node Encoder (multi-stream encoders) are shared with DeleteNet.

\paragraph{DeleteNet}
DeleteNet also builds off of the Node Encoder (off of the MLP outputs). In particular, for each node, it computes the difference between the Node Encoder output $\bb{v}_i - \bb{v}_{bg}$, where $\bb{v}_{bg}$ is the Node Encoder output of the background node. This difference is then passed through an MLP with 2 hidden layers of dimension 512 and output dimension of 1. The score is passed through a sigmoid to be in the range $[0,1]$. Weights of the Node Encoder are shared with SplitNet.

\subsection{Sampling a Split from SplitNet}
\label{subsec:splitnet_sampling}

We provide pseudocode of how to sample a split of mask $S \in \{0,1\}^{h \times w}$ given the output of SplitNet, which is a pixel-dense probability map $p \in [0,1]^{h \times w}$ in Algorithm~\ref{alg:split_sampling}. At a high level, we essentially compute a probability distribution over the contour of $S$ by seeing which pixels on the contour is close to split-able boundaries given by $p$ (more weight is given to split-able boundaries that are large components, since it is likely to find a path through that boundary). Then, start and end points are sampled and the lowest cost (where cost is 1 - $p$) is computed, scored and returned.

\begin{algorithm}[!t]
\caption{Sampling a Split}
\label{alg:split_sampling}
\begin{algorithmic}[1]
\REQUIRE Segmentation mask $S \in \{0,1\}^{h \times w}$, SplitNet output $p \in [0,1]^{h \times w}$, boundary threshold $\nu$.
\STATE Compute contour of $S$.
\STATE Threshold $p$ by $\nu$, and compute the connected components. Create an image $\tilde{p} \in \Z^{h \times w}$ where $\tilde{p}_i$ is the size (in pixels) of the component at $p_i$ for pixel $i$.
\STATE Compute contour probabilities for each contour pixel by weighted average of $\tilde{p}$ with Gaussian weights.
\STATE Sample start and end points on contour from contour probabilities.
\STATE Compute highest probability path from start to end through $p$, resulting in trajectory $\tau = \{(u_t, v_t)\}_{t=1}^{L}$.
\STATE Compute score $s_{\tau} = \frac{1}{L_i} \sum_t p_i[u_t, v_t]$.
\RETURN $\tau, s_{\tau}$
\end{algorithmic}
\end{algorithm}

\section{Segmentation Graph Scoring Network Details}
\label{sec:sgs_net_details}

A high-level illustration of SGS-Net can be found in Figure~\ref{fig:SGS-Net}.
The initial node features $\bb{v}_i^{(0)}$ are given by the Node Encoder, and we obtain initial edge features $\bb{e}_{ij}^{(0)}$ by running the Node Encoder on all neighboring union masks $S_{ij}$. Then, we run multiple Residual GraphNet Layers (RGLs). Our Residual GraphNet Layer is an adaptation of a GraphNet Layer~\cite{battaglia2018relational} with residual connections. Our RGL first applies an edge update: 
\begin{align}
    \bb{e}_{ij}^{(l+1)} &= \bb{e}_{ij}^{(l)} + \phi_e^{(l)}\left( \bb{v}_i^{(l)}, \bb{v}_j^{(l)}, \bb{e}_{ij}^{(l)} \right),
\end{align}
where $\phi_e^{(l)}$ is an MLP, and $l$ describes the layer depth. This is followed by a node update:
\begin{align}
    \mathcal{E}_i^{(l)} &= \left\{ \phi_{v_1}^{(l)}\left(\bb{e}_{ij}^{(l+1)}, \bb{v}_j^{(l)} \right) : (i,j) \in E \right\}\\
    \bb{v}_i^{(l+1)} &= \bb{v}_i^{(l)} + \phi_{v_2}^{(l)}\left( \overline{\mathcal{E}^{(l)}_i}, \bb{v}_i^{(l)} \right),
\end{align}
where $\overline{A}$ is the mean of all elements in the set $A$, and $\phi_{v_1}^{(l)}, \phi_{v_2}^{(l)}$ are MLPs. We additionaly apply ReLUs after the residual connections. After passing through $L$ levels of RGLs, we end up with the set of node and edge feature vectors $\mathcal{V} = \left \{\bb{v}_i^{(L)}\right\},\ \mathcal{E} = \left\{\bb{e}_{ij}^{(L)}\right\}$. We pass these through an output layer that aggregates these features:
\begin{equation}
    s_G = \sigma \left( \phi_o\left( \overline{\mathcal{V}}, \overline{\mathcal{E}} \right) \right) \in [0,1],
\end{equation}
where $\phi_o$ is yet another MLP and $s_G$ is the predicted graph score for segmentation graph $G$, and $\sigma$ is the sigmoid function.

\section{Implementation Details}
\label{sec:implementation_details}

Our Node Encoder is shared amongst all of the networks, including SplitNet, DeleteNet, and SGS-Net. We first jointly train the SO-Nets for 200k iterations, with one segmentation graph per network per iteration (the batch sizes is the number of instances in the segmentation graph) so that the Node Encoder contains useful information for both operations. Next, we hold the Node Encoder fixed while we train SGS-Net for 100k iterations. To train SGS-Net, we take an initial segmentation and perturb it with the four proposed sampling operations and compute their ground truth scores. However, we do not use the SO-Nets, instead we randomly split masks with sampled lines, merge neighboring masks, delete masks, and add masks in the same fashion as \citet{xie2019uois}. Modality tuning is performed during training of the SO-Nets, and held fixed during SGS-Net training. 

All images have resolution $H=480, W=640$. For our networks, we crop and resize the image, depth, and masks to $h=w=64$. All training procedures use Adam~\cite{kingma2015adam} with an initial learning rate of 1e-4. We use $K=3$ sample tree expansion iterations with a branching factor $B=3$. Max nodes and edges are set to $m_n=100, m_e=300$ during training, and $m_n=350, m_e=1750$ during inference. Undirected edges are handled by including both $(i,j)$ and $(j,i)$ as directed edges in the graph. For each segmentation graph, edges are connected between nodes if their corresponding masks are within 10 pixels in set distance. Additionally, when sampling a candidate graph, we first randomly choose a sampling operation, compute all possible perturbation scores (e.g. split scores $s_\tau$ for each mask), and randomly select 3 of these perturbations that have a score of 0.7 or higher. This gives the opportunity to explore more segmentations within the allotted budget. All experiments are trained and evaluated on a single NVIDIA RTX2080ti GPU.

\section{Pseudocode for Building the Sample Tree}

We provide pseudocode for building the sample tree in Algorithm~\ref{alg:sample_tree_CEM}.

\begin{algorithm}[t]
\caption{RICE}
\label{alg:sample_tree_CEM}
\begin{algorithmic}[1]
\REQUIRE Initial instance segmentation $S$, RGB image $I$, organized point cloud $D$.
\STATE Build $G_S$ with NodeEncoder applied to $I, D, S$
\STATE Initialize $T = \{G_S\}$
\FOR{$k \in [K]$}
\FOR{$G \in T\text{.leaves()}$}
\FOR{$b \in [B]$}
\STATE Randomly choose a sampling operation and apply it to $G$ to get candidate graph $G'$
\STATE Apply SGS-Net to obtain $s_{G'}, s_G$
\IF{$s_{G'} > s_G$}
\STATE Add $G'$ to $T$ as a child of $G$
\ENDIF
\IF{$T\text{.exceeds\_budget}(m_n, m_e)$}
\STATE Return $T$
\ENDIF
\ENDFOR
\ENDFOR
\ENDFOR
\RETURN Highest scoring graph in $T$ and/or contour uncertainties
\end{algorithmic}
\end{algorithm}

\section{Additional Experimental Results}

\subsection{Modality Tuning}
\label{subsec:modality_tuning}

\begin{figure*}[t]
     \centering
     \begin{subfigure}[b]{0.49\textwidth}
         \centering
         \includegraphics[width=\textwidth]{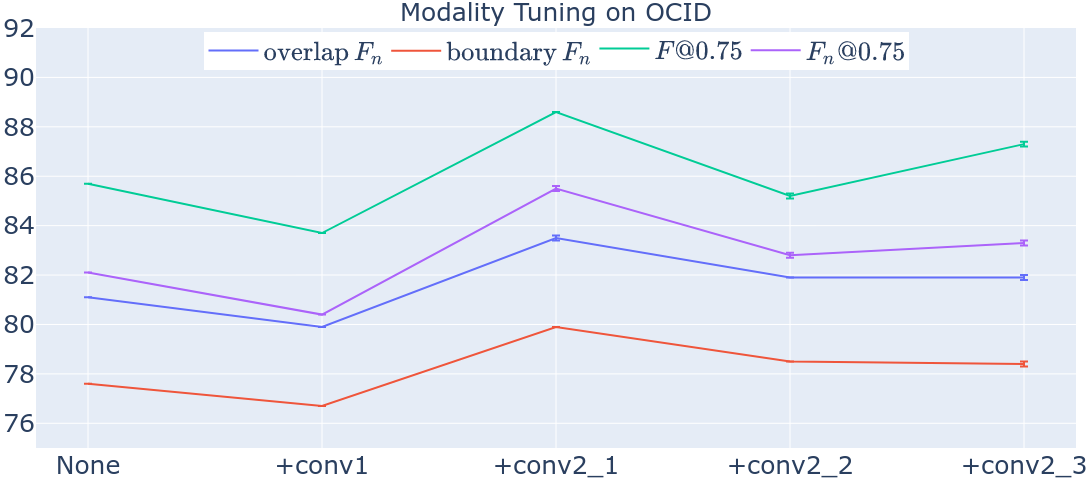}
     \end{subfigure}
     \hfill
     \begin{subfigure}[b]{0.49\textwidth}
         \centering
         \includegraphics[width=\textwidth]{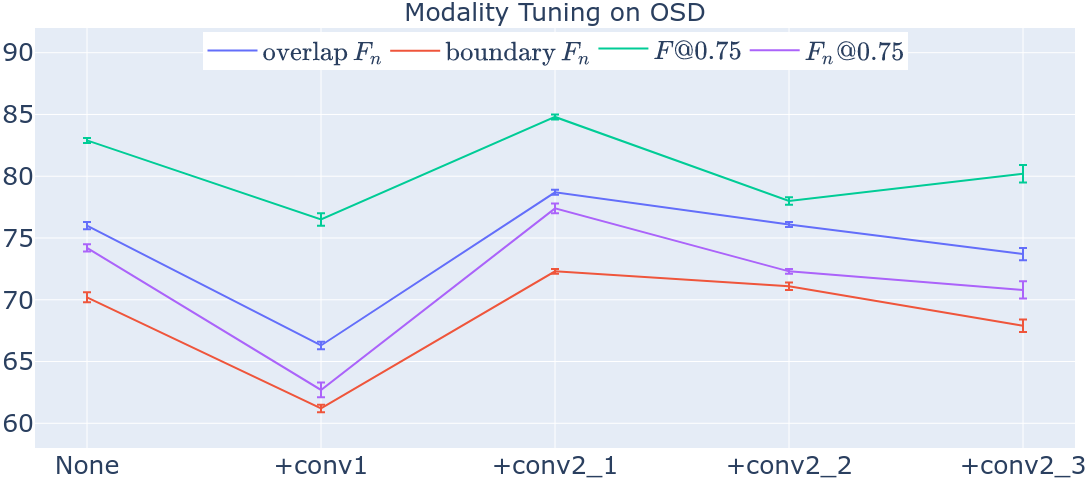}
     \end{subfigure}
     \caption{Modality tuning on OCID~\cite{suchi2019easylabel} and OSD~\cite{richtsfeld2012segmentation} shows that tuning up to conv2\_1 when training on simulated data generalizes best to real data. Note that standard deviation bars are shown, but are very tight and difficult to see.}
     \label{fig:modality_tuning}
\end{figure*}

Inspired by \citet{aytar2017cross}, we perform an ablation to study how to best generalize from our non-photorealistic dataset TOD to real-world data. Starting from a ResNet50 pre-trained on COCO~\cite{lin2014microsoft}, we ablate over tuning the \texttt{conv1} layer, the \texttt{conv2\_1}, \texttt{conv2\_2}, \texttt{conv2\_3} bottleneck building blocks~\cite{he2016deep}, or keeping ResNet fixed. The idea is that by fixing the rest of the layers, we can encourage ResNet to learn the high level representation it has learned on COCO, on our simulated dataset. Thus, our SO-Nets and SGS-Net will learn to consume this high-level representation to provide their predictions. Then, during inference in the real-world, we resort back to the pre-trained ResNet to extract that representation from real images. 
In Figure~\ref{fig:modality_tuning}, we show the results of our experiment. Interestingly, only modality-tuning \texttt{conv1} leads a small dip in performance compared to no tuning, while the optimal tuning for our scenario is to tune \texttt{conv1} and \texttt{conv2\_1}. For the rest of this section, all networks will have been trained with this optimal setting.

\subsection{Evaluating the Usefulness of Each Sampling Operation}
\label{subsec:so_ablation}

We provide an ablation experiment where we test the efficacy of each sampling operation. In particular, we test RICE while only using one sampling operation at a time, so that the sample tree is built only by a particular operation, e.g. splitting. This allows us to determine which of the sampling operations is most helpful in comparison to the initial instance segmentation method (e.g. UOIS-Net-3D~\cite{xie2021unseen}, UCN~\cite{xiang2020learning}). 

We test \textbf{Split} only, \textbf{Merge} only, and \textbf{Delete/Add} only. Note that we group Delete and Add together since the Add operation is essentially the Delete operation after extracting connected components from an external foreground mask $F$ (See Section 3.3). In Table~\ref{table:sample_operation_ablation}, we show the results for $F$@.75 and $F_n$@.75 metrics using UCN~\cite{xiang2020learning}, UOIS-Net-3D~\cite{xie2021unseen}, Mask R-CNN~\cite{he2017mask}, and PointGroup~\cite{jiang2020pointgroup} as initial instance segmentation methods. We also show the percentage of increased performance with respect to the increased performance when using all sampling operations in parentheses. Clearly, we see that the split operation alone results in most of the performance gain compared to the full RICE method. This indicates that all four initial instance segmentation methods tend to under-segment, which is a common failure case in densely cluttered environments. UCN gains a lot from merging; the reason for this is that a common failure case from their pixel-clustering procedure is that the boundaries of the objects tend to be clustered as a separate object whichh results in over-segmentation. Merging can easily solve this issue. Lastly, Mask R-CNN and PointGroup also benefit from delete/add, which suggests that they are either predicting false positives and/or false negatives.

\begin{table*}[t]
\centering
\resizebox{\linewidth}{!}{\begin{tabular}{c||cccccc}
\toprule 
Initial Instance & \multicolumn{2}{c}{\textbf{Split} only} & \multicolumn{2}{c}{\textbf{Merge} only} & \multicolumn{2}{c}{\textbf{Delete/Add} only} \\
Segmentation Method & \Fsf & \Fnsf & \Fsf & \Fnsf & \Fsf & \Fnsf \\
\midrule
UCN~\cite{xiang2020learning} & 92.0 \relpc{100} & 87.2 \relpc{64.7} & 88.8 (\textcolor{red}{-23.1\%}) & 87.5 \relpc{73.5} & 89.5 \relpc{3.8} & 86.9 \relpc{55.9} \\
UOIS-Net-3D~\cite{xie2021unseen} & 88.5 \relpc{101} & 84.6 \relpc{91.6} & 78.4 \relpc{2.1} & 77.4 \relpc{4.8} & 78.3 \relpc{1.1} & 77.1 \relpc{1.2} \\
Mask R-CNN~\cite{he2017mask} & 79.7 \relpc{60.3} & 75.8 \relpc{82.0} & 66.0 \relpc{0.0} & 64.6 \relpc{1.4} & 69.1 \relpc{13.6} & 67.2 \relpc{20.1} \\
PointGroup~\cite{jiang2020pointgroup} & 82.4 \relpc{86.1} & 77.4 \relpc{87.7} & 61.2 \relpc{1.6} & 60.7 \relpc{2.1} & 64.1 \relpc{13.1} & 63.1 \relpc{14.4} \\
\bottomrule
\end{tabular}}
\caption{Sampling Operation Ablation. We omit standard deviations as they are all less than 0.0005. We show results on $F$@.75 and $F_n$@.75. In parentheses, we show relative gain compared to the full RICE method (with all sampling operations). }
\label{table:sample_operation_ablation}
\end{table*}

\subsection{Full Results on P/R/F}

\begin{figure*}[t]
     \centering

     \begin{subfigure}[b]{0.8\textwidth}
         \centering
         \includegraphics[width=\linewidth]{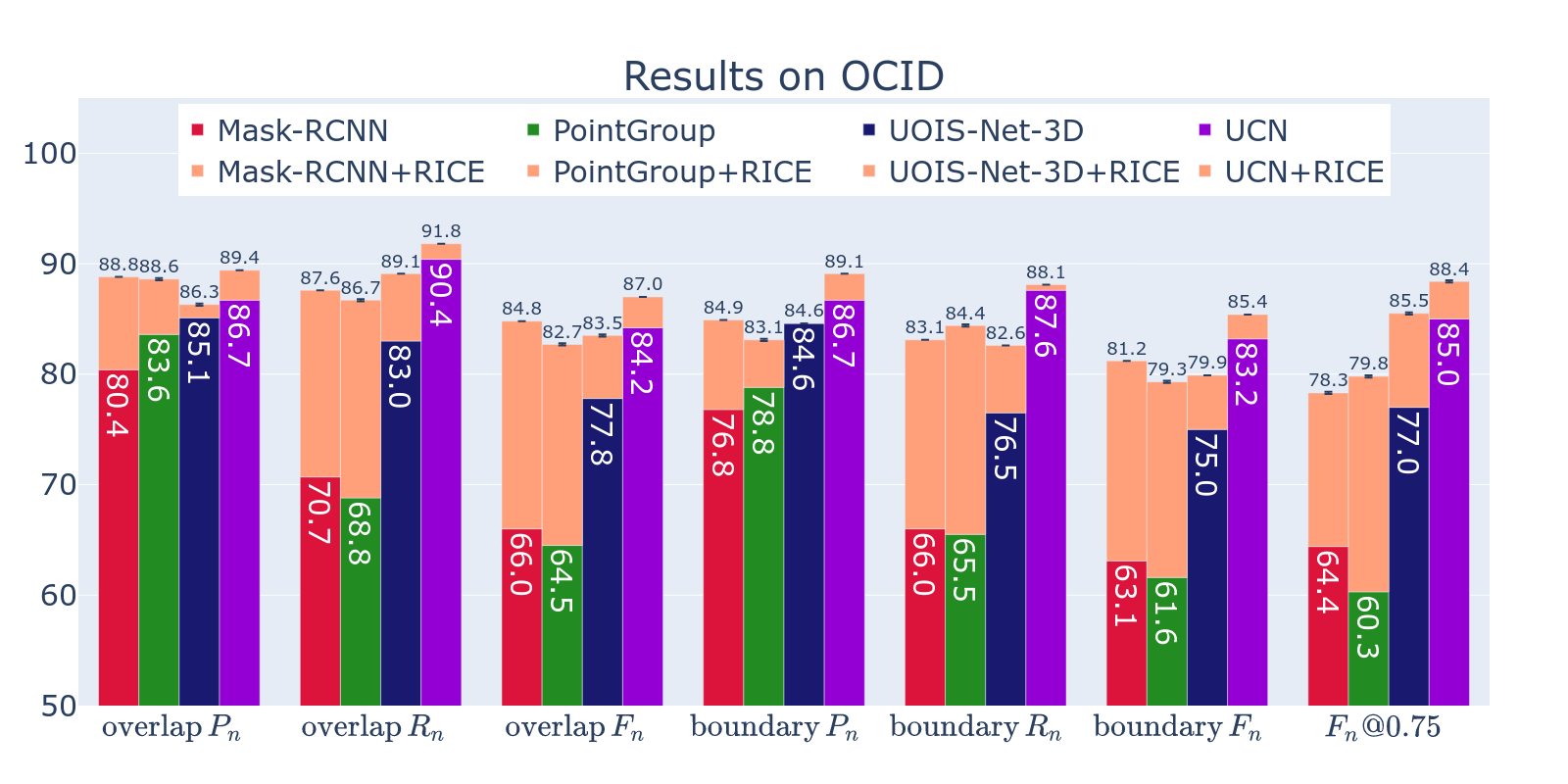}
         \caption{Results on OCID~\cite{suchi2019easylabel} using OSN metrics.}
         \label{fig:SOTA_improvement_OCID_osn_full}
         \vspace{5mm}
     \end{subfigure}

     \begin{subfigure}[b]{0.8\textwidth}
         \centering
         \includegraphics[width=\linewidth]{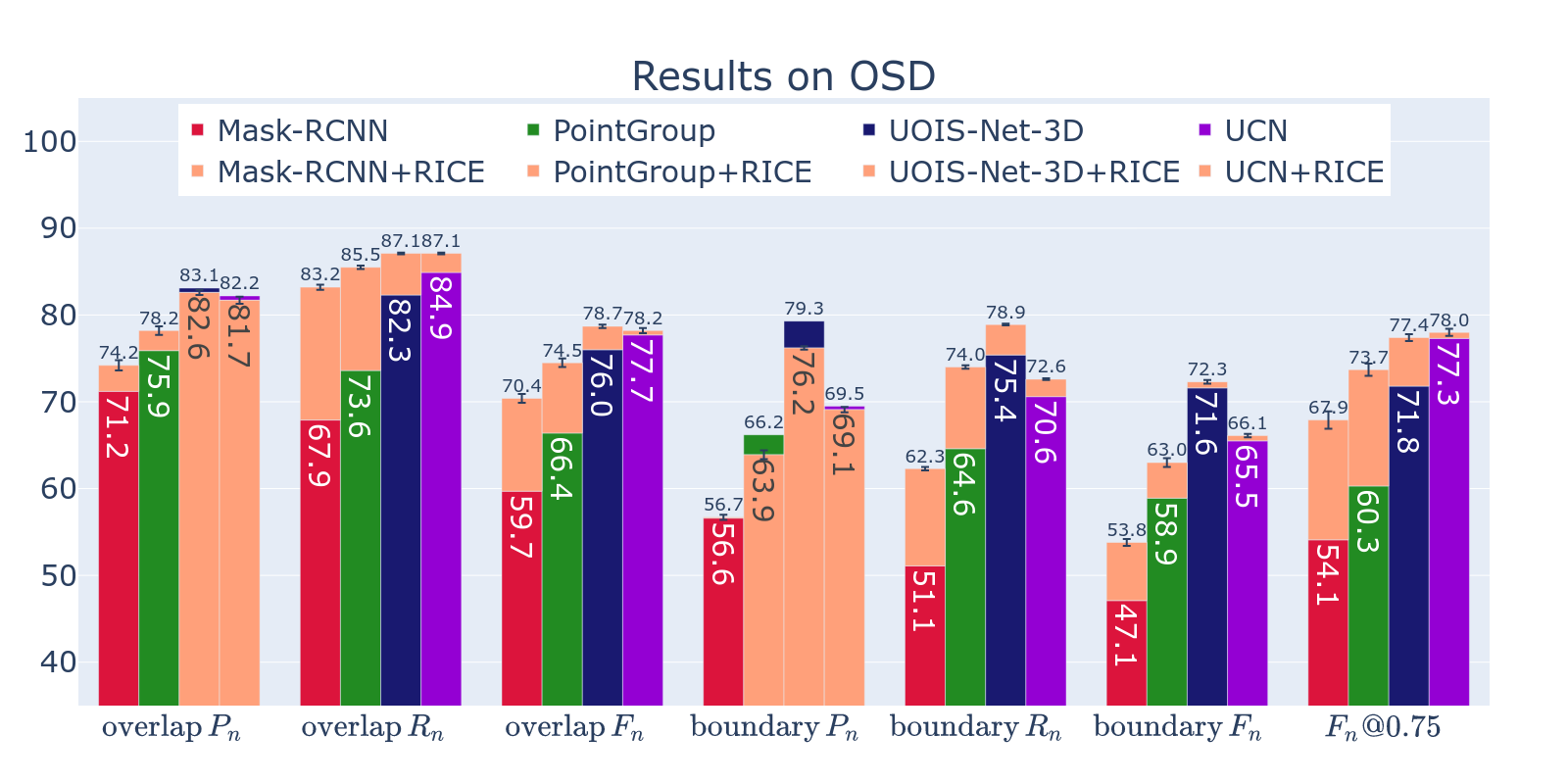}
         \caption{Results on OSD~\cite{richtsfeld2012segmentation} using OSN metrics.}
         \label{fig:SOTA_improvement_OSD_osn_full}
     \end{subfigure}
\end{figure*}

We provide full results on all Object Size Normalized (OSN) metrics for OCID~\cite{suchi2019easylabel} and OSD~\cite{richtsfeld2012segmentation} in Figures \ref{fig:SOTA_improvement_OCID_osn_full} and \ref{fig:SOTA_improvement_OSD_osn_full}. For OCID, we can see increases in performance across all metrics in light orange. When RICE underperforms the initial segmentation, we color the bar underneath as light orange. On OSD, we can see that overlap $P_n$ and boundary $P_n$ are slightly worse than the initial method, while the recall $R_n$ and $F_n$ measures are still higher. 

\begin{figure*}[t]
     \centering
    
     \begin{subfigure}[b]{0.8\textwidth}
         \centering
         \includegraphics[width=\linewidth]{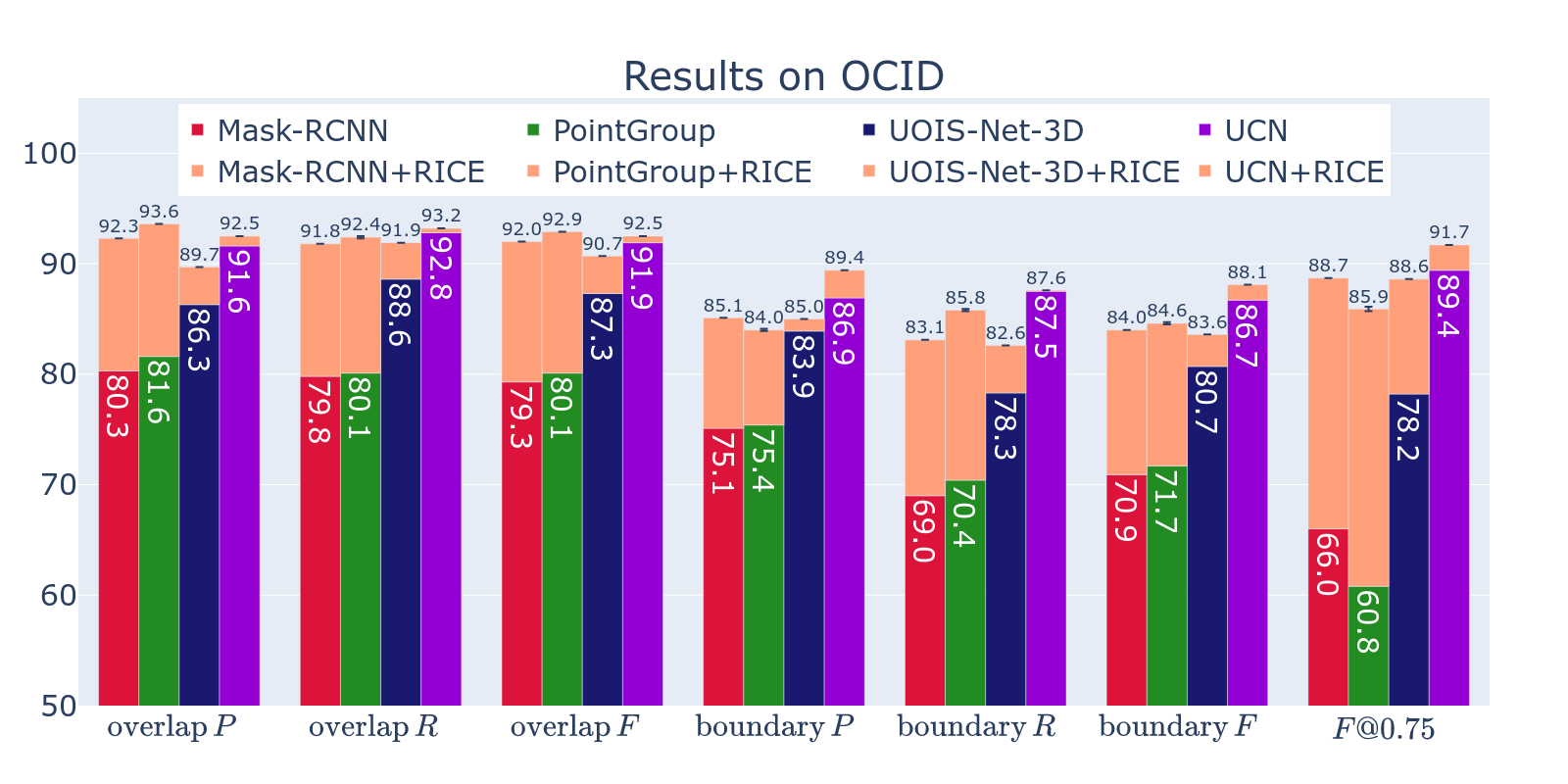}
         \caption{Results on OCID~\cite{suchi2019easylabel} using Overlap and Boundary P/R/F metrics as defined by~\citet{xie2019uois}. Note that the UCN~\cite{xiang2020learning} numbers are slightly different than in the published paper, due to the authors finding a bug in the reporting code. This is the case for \citet{xie2021unseen} as well.}
         \label{fig:SOTA_improvement_OCID_normal_full}
         \vspace{5mm}
     \end{subfigure}

     \begin{subfigure}[b]{0.8\textwidth}
         \centering
         \includegraphics[width=\linewidth]{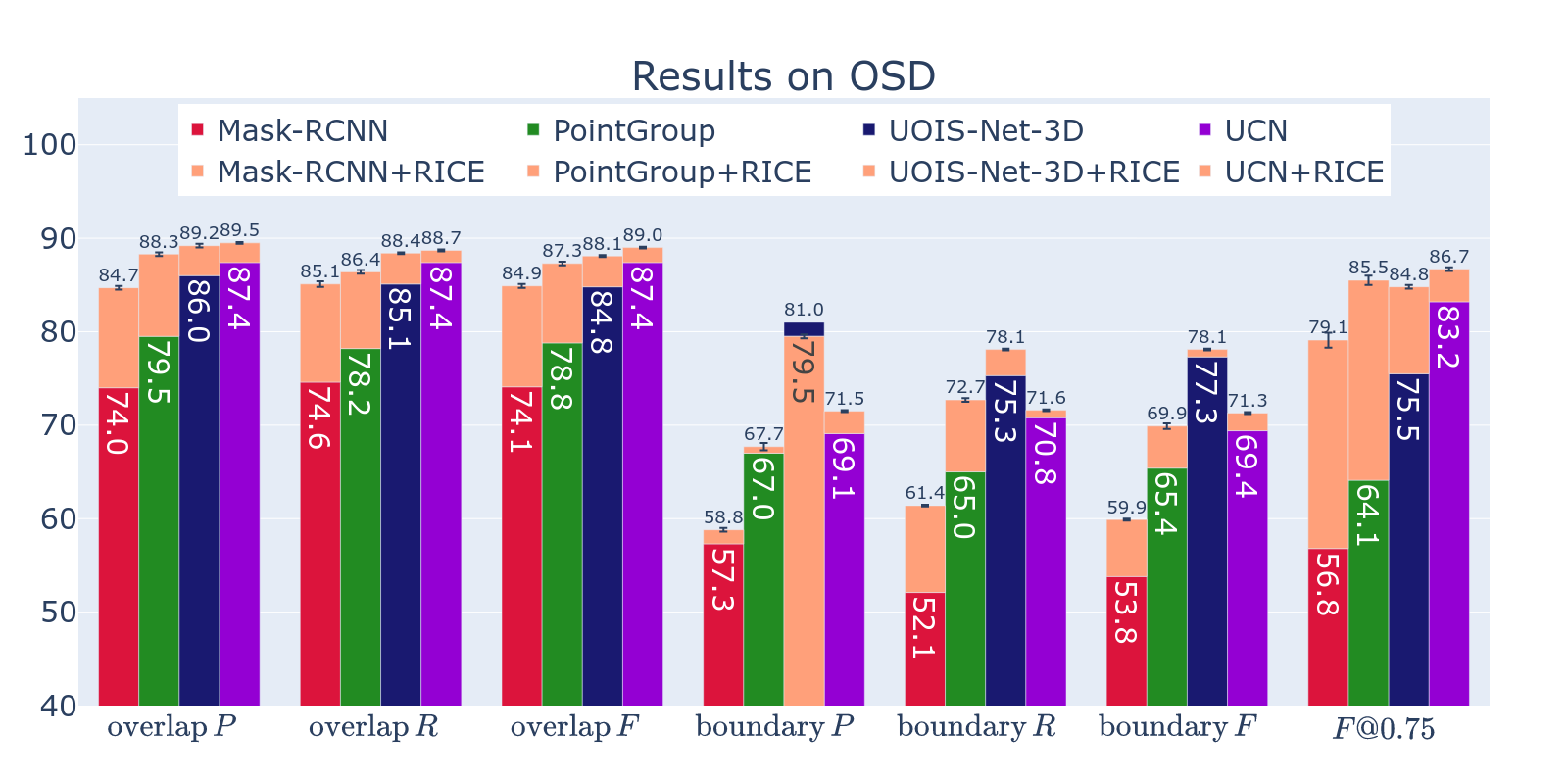}
         \caption{Results on OSD~\cite{richtsfeld2012segmentation} using Overlap and Boundary P/R/F metrics as defined by~\citet{xie2019uois}.}
         \label{fig:SOTA_improvement_OSD_normal_full}
     \end{subfigure}
     
\end{figure*}

Figures \ref{fig:SOTA_improvement_OCID_normal_full} and \ref{fig:SOTA_improvement_OSD_normal_full} show Overlap and Boundary P/R/F measures on OCID and OSD. They show similar results to the OSN measures, but the numbers are higher. This suggests that common mistakes for SOTA methods more commonly occur in smaller objects (investigations of failure cases from \citet{xie2021unseen, xiang2020learning} suggest this is the case). These numbers are directly comparable with previously published works.

\section{More Details for: Guiding a Manipulator with Contour Uncertainties for Efficient Scene Understanding}


To extract contour uncertainties, we exploit the fact that RICE is a stochastic algorithm by design. Each leaf is essentially a sampled trajectory of states (segmentations) and actions (perturbations) from the initial segmentation $S$. These trajectories may have explored different parts of the state space (e.g. perturbed different object masks in the scene). We compute the standard deviation of the contours of each leaf graph in order to provide contour uncertainty estimates, which are shown in red in Figure~\ref{fig:uncertainty_grasp} and the Supplemental video.
Additionally, we visualize the confident contours (which are present in each leaf graph) in green.

The contour uncertainties depict certain segmentations that not all of the trajectories explored. It reflects which objects RICE is not as confident about. If a mask $S_i$ in the initial segmentation is split the same way in all leaf graphs, then RICE is confident that $S_i$ should be split, and there will be no uncertainty. However, if $S_i$ is only split in some of the leaf graphs, then RICE is not as confident about whether $S_i$ truly represents more than 1 object, and an interaction is required to resolve such uncertainty. For example, in Trial 2 in the Supplemental video, the cup and the bowl it is on top of (far left) are constantly under-segmented together by UCN~\cite{xiang2020learning}. RICE splits it correctly each time, and there is no uncertainty about splitting that mask, as evidenced by the uncertainty contours.

We provide a short description of the grasping algorithm with pseudocode in Algorithm~\ref{alg:grasping_for_uncertainty}.

\begin{algorithm}[t]
\caption{Guiding a Manipulator with RICE}
\label{alg:grasping_for_uncertainty}
\begin{algorithmic}[1]
\REPEAT
\STATE Get $S$ from initial instance segmentation method (e.g. UCN~\cite{xiang2020learning})
\STATE Run RICE to get best masks and contour uncertainty
\IF{Contour Uncertainty is present}
\STATE Sample a grasp with~\citet{sundermeyer2021icra} from the uncertain masks, and execute it
\ENDIF
\UNTIL{No Contour Uncertainty}
\end{algorithmic}
\end{algorithm}

\section{Example Sample Tree}

\begin{figure*}[!h]
    \centering
    \includegraphics[width=\linewidth]{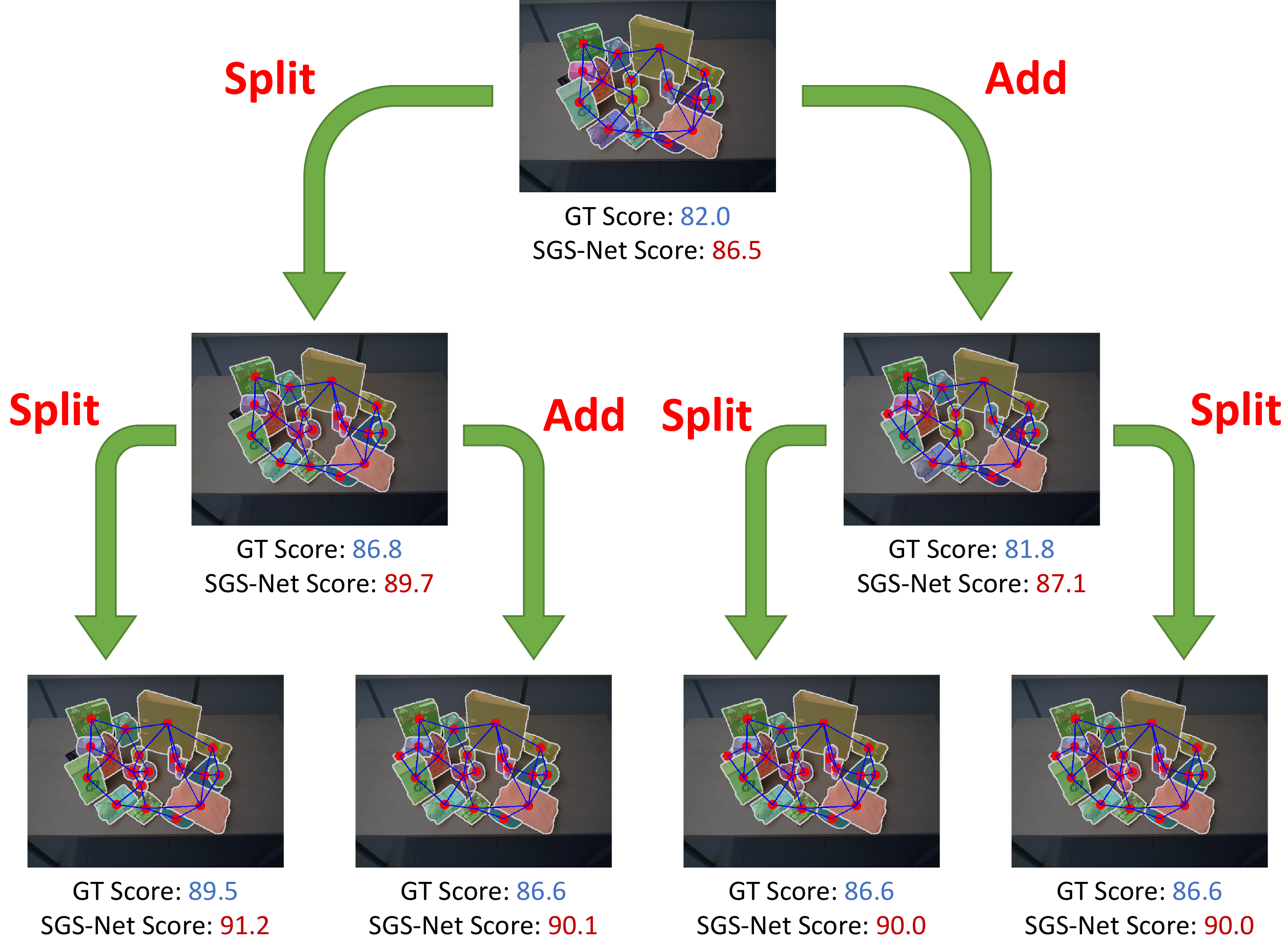}
    \caption{Example of a sample tree. Ground Truth and SGS-Net scores are shown, along with the chosen sampling operations. In this example, all leaves improve upon the initial segmentation graph, with the highest ranking graph also being the closest to the ground truth segmentation. Very similar splits and adds are investigated in the leaf trajectories.}
    \label{fig:sample_tree_example}
\end{figure*}

In Figure~\ref{fig:sample_tree_example}, we show an example of a sample tree with branch factor $B=2$ and $K=2$ expansion iterations. We also visualize the ground truth score ($.8 F + .2  F\text{@.75}$) and the predicted score from SGS-Net. Note that the SGS-Net scores improve as the graph node gets further away from the root. In this particular example, SGS-Net would return the bottom left graph, which is also the most accurate graph.

\end{document}